\documentclass[twoside,11pt]{article}
\usepackage{mathptmx}
\usepackage{jmlr2e}
\hypersetup{hidelinks} 
\usepackage{amsmath,mathtools}
\usepackage{booktabs,microtype,enumitem}
\newtheorem{assumption}[theorem]{Assumption}

\newcommand{\Pmix}{P_{\mathrm{mix}}}
\newcommand{\Pid}{P_{\mathrm{ID}}}
\newcommand{\Pood}{P_{\mathrm{OOD}}}
\newcommand{\Fmix}{F_{\mathrm{mix}}}
\newcommand{\Fid}{F_{\mathrm{ID}}}
\newcommand{\Food}{F_{\mathrm{OOD}}}
\newcommand{\eps}{\varepsilon}
\newcommand{\ind}[1]{\mathbf{1}\{#1\}}
\newcommand{\cmark}{\checkmark}
\newcommand{\xmark}{\ensuremath{\times}}

\newcommand{\CONFcollapseOODD}{$-0.114$}
\newcommand{\CONFcollapseADA}{$-0.028$}

\jmlrheading{}{2026}{}{}{}{}{Vishnu Bindu Balachandran}
\ShortHeadings{Self-Poisoning in Adaptive OOD Detection}{Vishnu Bindu Balachandran}
\firstpageno{1}

\begin{document}

\title{Self-Poisoning in Adaptive Out-of-Distribution Detection:\\
A Sharp-Threshold Theory and Certified Label-Free Calibration}

\author{\name Vishnu Bindu Balachandran
        \email vishnubindubalachandran@outlook.com \\
       \addr Independent Researcher \\
       \addr www.vishnubindubalachandran.com}

\editor{}

\maketitle

\begin{abstract}%
Test-time adaptive out-of-distribution (OOD) detectors update a memory bank from the
unlabelled stream. We show this adaptation obeys a provable dynamical law. Modelling
bank impurity as a generalized P\'olya urn, we prove almost-sure convergence to a
mean-field equilibrium whose slope acts as a reproduction number. Below one, impurity
stays benign. Above one, the bank is fully poisoned and the detector collapses. The
measured admission kernel is affine ($R^2 \ge 0.996$) with slope just below one in every
encoder family (a protocol signature), so this detector class is near-critical by
design, and across 96
settings the predicted threshold matches the empirical collapse, where ungated
dictionaries lose up to $0.163$ AUROC. We then prove that a certified admission gate,
reading only a frozen reserve, severs the feedback loop and removes the transition at
every contamination rate, even adversarially, while controlling false positives
label-free. For the complementary static-calibration failure under drift we give CDC,
which restores nominal FPR label-free on all tested drift-affected cells. Finally we
prove a two-world impossibility theorem. Drift and contamination are indistinguishable
without labels, forcing a closed-form power ceiling our procedure approaches. Together
these give a complete possibility/impossibility characterization of label-free adaptive
OOD detection.
\end{abstract}


\section{Introduction}\label{sec:intro}

A deployed OOD detector faces two requirements that pull in opposite directions. It must
control false positives at a nominal rate $\alpha$ (every flagged input incurs
review cost, and a detector whose false-positive rate (FPR) silently doubles is not
usable in production), and it must adapt, because both the in-distribution (ID)
data and the outliers it sees at test time differ from anything available at training.
The modern test-time-adaptation literature resolves this tension by pseudo-labelling:
memory-bank detectors append the most ID-looking stream points to an ID bank
\citep[e.g.][AdaODD]{adaodd}, or the most OOD-looking points to an OOD dictionary
\citep[e.g.][OODD]{oodd}, and re-score against the updated banks. These methods report
substantial gains under favorable stream mixes. They also, as we show, fail in a
specific, predictable, and severe way under realistic ones.

\begin{figure}[t]
\centering
\includegraphics[width=\textwidth]{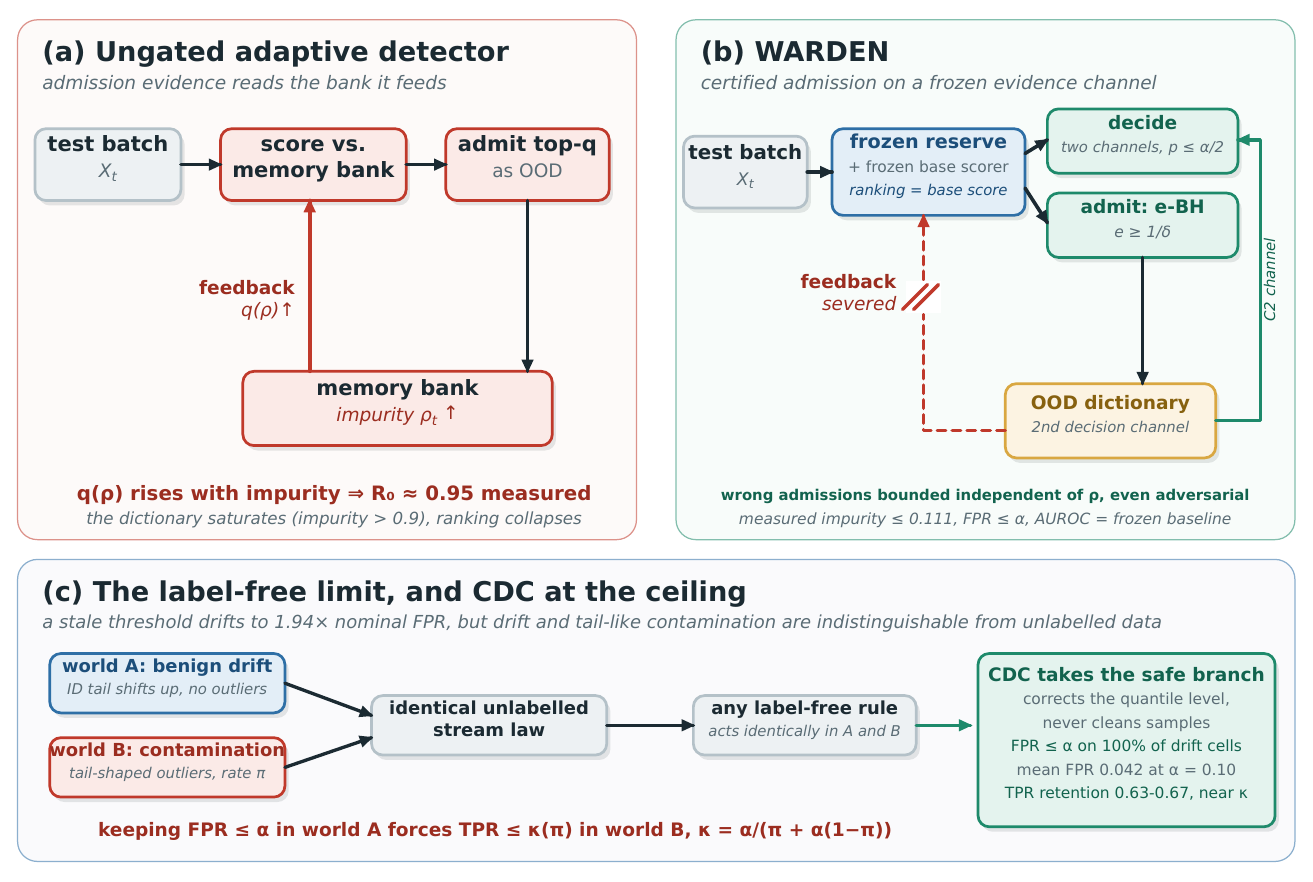}
\caption{The paper in one figure. (a)~Ungated adaptive detectors compute admission
evidence against the very bank that evidence feeds, so the false-admission kernel
$q(\rho)$ rises with impurity. The aggregate slope sits just below one
($R_0\approx0.95$, a signature of the fixed-fraction admission protocol,
Appendix~\ref{app:exp}), the dictionary saturates (impurity $>0.9$), and ranking
collapses (Sections~\ref{sec:setting} and~\ref{sec:theory}). (b)~WARDEN severs the loop.
Admission evidence is conformal against a frozen reserve under the frozen base
scorer, e-BH requires $e\ge1/\delta$ of every admission, and the dictionary feeds
only a second decision channel, never the admission evidence (that channel does not
improve detection on our features, Appendix~\ref{app:c2}; the severed admission is
what matters). The wrong-admission intensity is therefore
bounded independently of the dictionary state, even under adversarially chosen
contamination (Lemma~\ref{lem:ebh}, Theorem~\ref{thm:gating},
Corollary~\ref{cor:adversarial}). Decisions keep FPR $\le\alpha$ by conformal
validity (Proposition~\ref{prop:validity}). Measured impurity never exceeds $0.111$,
and AUROC tracks the frozen baseline by construction. (c)~The complementary failure
and its limit. A stale threshold drifts to $1.94\times$ nominal FPR, yet benign
drift and tail-like contamination generate identical unlabelled streams, so any
label-free rule keeping FPR $\le\alpha$ under drift caps its power under
contamination at $\kappa(\pi)$ (Theorem~\ref{thm:impossible}). CDC corrects the
calibration quantile on the contaminated stream itself, certifies FPR $\le\alpha$
on all tested drift-affected cells, and retains a median $0.63$--$0.67$ of oracle
power, near the ceiling (Section~\ref{sec:cdc}).}
\label{fig:arch}
\end{figure}

\paragraph{Self-poisoning is a phase transition, not noise.}
When the stream is temporally clustered (bursty) and contamination is low (the
standard regime of production monitoring), a pure-ID batch forces an ungated OOD
dictionary to admit ID tail points. The contrast score of nearby ID points then inflates,
which makes further ID admissions more likely (Figure~\ref{fig:arch}a). That this loop can hurt is folklore in the test-time-adaptation literature. What has
been missing is a prediction of when. We prove the loop is a reinforced stochastic
process whose impurity
$\rho_t$ (fraction of wrongly admitted points) converges almost surely to the stable
equilibria of a scalar mean field $h(\rho)=q(\rho)-\rho$, where $q$ is the admission
kernel, the probability that the next admitted point is wrong given current impurity
(Section~\ref{sec:theory}). For the affine kernels we measure across all our settings,
the slope $b=q'(\rho)$ acts as a reproduction number. When $b<1$ the equilibrium is benign,
$\rho^*=a/(1-b)$. When $b\ge 1$ poisoning is complete, $\rho_t\to 1$. Saturating kernels
produce a fold bifurcation, a discontinuous jump of the equilibrium at a critical
contamination rate $\pi_c$, with hysteresis. (Our protocol
probes the threshold and knee, not the hysteresis, which would require $\pi$-sweeps within a
run.) Empirically, the predicted $\pi_c$ matches
the observed knee on all 96 settings we measure, and ungated dictionaries collapse
to $0.163$ AUROC below their own frozen baseline in the bursty low-contamination regime, below chance on near-OOD pairs, while their realized FPR departs arbitrarily from
the nominal level.

\paragraph{Certified admission removes the transition.}
The same theory yields the fix and pinpoints why it works. The collapse loop
exists because admission evidence is computed against the very bank it feeds. WARDEN
severs the loop (Figure~\ref{fig:arch}b). Its admission e-values are conformal against a frozen reserve,
never the dictionary, and it gates them with e-BH \citep{wang2022ebh} at level
$\delta$. Every admission must clear the absolute threshold $e \ge 1/\delta$, so the
wrong-admission intensity is bounded by a constant independent of the dictionary
state. The reinforcement term is structurally absent and the supercritical branch
disappears for every contamination rate (Theorem~\ref{thm:gating}), while e-BH keeps
each batch's expected false-admission proportion at $\delta$ under arbitrary
dependence. No admission evidence depends on anything the contamination can influence,
so these guarantees hold even when the contaminated points are chosen by an adaptive
adversary (Corollary~\ref{cor:adversarial}). The resulting detector tracks its frozen
baseline AUROC exactly (by construction, it ranks with the base score, a design
invariant rather than a performance claim), holds mean realized FPR at $0.056$
(per-family means $0.047$--$0.061$; per-cell $5$--$95\%$ range $0.036$--$0.100$, max
$0.121$) across the $4{,}800$ standard-configuration cells of the kill, confirmation,
and grid campaigns (the ablation campaign varies WARDEN's own knobs and is reported
separately), and its measured dictionary impurity never exceeds $0.111$ there, against
ungated impurity above $0.9$.

\paragraph{Static calibration fails too, and fixing it label-free has a price.}
The frozen alternative is not safe either. A detector calibrated on train-ID data (all a deployment has) realizes FPR $1.94\times$ nominal across our settings, because
of train$\to$test ID drift (a per-dimension variance drift in whitened feature space; a
mean-shift correction recovers nothing). We give CDC (Section~\ref{sec:cdc}). It calibrates
the operating threshold directly on the contaminated stream at a contamination-corrected
quantile level $1-\alpha(1-\hat\pi_{\mathrm{up}})$ plus a DKW slack, where
$\hat\pi_{\mathrm{up}}$ is a Storey-type upper confidence estimate computed from
conformal p-values against the stale reserve. The drift that breaks the stale
calibration biases the Storey statistic upward, in exactly the conservative
direction, and we give a sufficient condition under which it remains a valid upper
bound (Theorem~\ref{thm:storey}). CDC needs no labels at any point, certifies
FPR~$\le\alpha$ with high probability (Theorem~\ref{thm:cdc}), and empirically certifies
all $1{,}695$ tested drift-affected cells at mean FPR $0.042$, retaining a median
$67\%$ of oracle TPR.

\paragraph{The retained power cannot be pushed to $100\%$.}
Our impossibility result (Theorem~\ref{thm:impossible}) constructs two worlds (benign
ID drift versus contamination by outliers distributed exactly like the ID tail) whose
unlabelled stream laws are identical (Figure~\ref{fig:arch}c). Any label-free procedure behaves identically
in both, so if it maintains FPR $\le\alpha$ under the drift world its power in the
contamination world is capped at $\kappa=\alpha/(\pi+\alpha(1-\pi))<1$. Label-free
adaptive calibration therefore must pay a power price set by the confusable
contamination mass. CDC's measured power retention ($0.67$; $0.63$ held-out) sits near
this worst-case ceiling ($0.69$ at $\pi{=}0.05$), which bounds what any
certificate-preserving label-free method could add on adversarial instances, though
on benign instances the ceiling is higher and part of CDC's gap is finite-sample slack
(Remark~\ref{rem:optimal} states the comparison precisely).

\paragraph{Contributions.}
\begin{enumerate}[leftmargin=1.4em,itemsep=1pt,topsep=2pt]
\item \textbf{A sharp-threshold theory of self-poisoning} (Section~\ref{sec:theory}):
almost-sure convergence of dictionary impurity to mean-field equilibria
(Theorem~\ref{thm:converge}); reproduction number and critical contamination for affine
kernels (Corollary~\ref{cor:affine}); fold bifurcation and hysteresis for saturating
kernels (Proposition~\ref{prop:fold}); finite-window behavior under FIFO eviction
(Proposition~\ref{prop:fifo}); and structural removal of the transition by
severed-feedback certified admission, robust to adaptive adversaries
(Lemma~\ref{lem:ebh}, Theorem~\ref{thm:gating}, Corollary~\ref{cor:adversarial}).
\item \textbf{CDC: certified decontaminated calibration} (Section~\ref{sec:cdc}): the
first label-free procedure that restores nominal FPR under simultaneous ID drift and
stream contamination, with finite-sample validity (Theorem~\ref{thm:cdc}), a
drift-conservativity guarantee for its contamination estimate (Theorem~\ref{thm:storey}),
and an explicit power bound (Theorem~\ref{thm:power}).
\item \textbf{A matching impossibility theorem} (Section~\ref{sec:impossibility}):
without separation assumptions, no label-free procedure can simultaneously control FPR
under drift and retain power under contamination. The assumption CDC uses is necessary,
and its power loss is unavoidable in order of magnitude.
\item \textbf{The largest cross-family measurement of adaptive OOD dynamics to date}:
96 settings $\times$ contamination $\times$ ordering $\times$ seeds ($9{,}264$ streaming
cells over the primary campaigns, plus a $3{,}840$-cell one-shot held-out replication
that confirms every headline claim).
\end{enumerate}

\paragraph{What we do not claim.} Adaptation does not improve ranking quality on
well-whitened features: growing the ID bank with thousands of true-ID points moves AUROC
by $<0.005$ on our settings, and OOD-dictionary contrast scores hurt when clean. The
certified dictionary decision channel does not improve detection at a fixed FPR budget
either (Appendix~\ref{app:c2}). We therefore make no state-of-the-art AUROC claim
anywhere. The contributions are the
dynamical law, the certificates, and the impossibility boundary, the quantities that
decide whether an adaptive detector is deployable at all.

\section{Related work}\label{sec:related}

\paragraph{Test-time adaptive OOD detection and its failures.}
Memory-bank and dictionary methods (AdaODD \citep{adaodd}, OODD \citep{oodd}, and the
broader test-time-adaptation line) adapt from unlabelled streams via pseudo-labels.
Training-time approaches instead exploit unlabelled mixtures before deployment and
require retraining \citep{du2024sal, woods2022}.
Collapse of self-training under noisy pseudo-labels is documented qualitatively (error accumulation and model collapse in long-term TTA \citep{ttacollapse,zsntta}, and
adversarial test-time poisoning \citep{ttapoison}), but, to our knowledge, no prior
work identifies the phenomenon as a reinforced stochastic process with a provable
critical threshold, nor derives the gate that removes it. Robust-TTA heuristics
(sample-selection, resets, anti-forgetting) lack guarantees by construction.
Table~\ref{tab:guarantees} places the present paper in this landscape. The test-time
poisoning attacks of \citet{ttapoison} operate through the mutable state their targets
consult at decision time. Corollary~\ref{cor:adversarial} shows that certified
admission removes this attack surface by construction rather than by hardening.
The defense philosophy, constrain what the stream may admit so that poisoning is
bounded, goes back to \citet{kloft2012security}, who bound the displacement an
adversary can force on an online centroid anomaly detector under a false-positive
budget. WARDEN can be read as a conformal completion of that program: its admission
budget is enforced by e-BH against a frozen reserve, which makes the bound label-free
and independent of the bank state and of the contamination process.
At training scale, self-consuming loops now have a developed theory. Models degrade
when trained on their own outputs \citep{shumailov2024collapse}, iterative retraining
is stable when the clean-data fraction is large enough \citep{bertrand2024stability},
accumulating rather than replacing data bounds the error
\citep{gerstgrasser2024collapse}, and in high-dimensional regression even a vanishing
synthetic fraction can preclude consistency \citep{dohmatob2025strong}. Our
subcritical branch does not contradict the last result. Strong collapse concerns the
asymptotic bias of retraining model weights on unboundedly accumulating synthetic
data, whereas Corollary~\ref{cor:affine} concerns the stationary composition of a
capped, evicting bank whose per-step influence is damped, and a benign impurity
equilibrium is a statement about the bank, not a claim that contamination is harmless
downstream. What this paper adds to that conversation is a deployed-detector setting
where the loop admits an exact scalar mean field with a measurable kernel, a
certified gate that provably removes the transition, and a matching impossibility
bound.

\begin{table}[t]
\centering
\caption{Guarantee landscape. A poisoning bound is a bank-impurity bound that holds
for every contamination rate and stream ordering. FPR under drift means a finite-sample
false-positive certificate on the drifted stream without labels. \cmark{} and \xmark{}
mark whether a guarantee is provided, a dash marks a guarantee that does not apply
because the method keeps no adaptive bank, and a theorem number marks where we prove it.
For FPR under drift we also note each prior method's binding restriction.}
\label{tab:guarantees}
\small
\setlength{\tabcolsep}{5pt}
\resizebox{\textwidth}{!}{%
\begin{tabular}{lccccc}
\toprule
method & adapts & label-free & poisoning bound & FPR under drift & power bound \\
\midrule
AdaODD, OODD & bank & \cmark & \xmark & \xmark & \xmark \\
robust-TTA heuristics & model, bank & \cmark & \xmark & \xmark & \xmark \\
\citet{bashari2025} & \xmark & \xmark & -- & fixed reference & \cmark \\
\citet{wang2026trimming} & \xmark & \cmark & -- & no drift & \xmark \\
QTC \citep{yilmaz2022qtc} & threshold & \cmark & -- & no contamination & \xmark \\
this paper & bank + threshold & \cmark & Thm~\ref{thm:gating} & Thm~\ref{thm:cdc} & Thm~\ref{thm:power} \\
\bottomrule
\end{tabular}}
\end{table}

\paragraph{Conformal novelty detection and e-values.}
Conformal p-values for outlier detection, FDR-controlling selection, and e-value
calibration are established \citep{bates2023testing, wang2022ebh, jin2023selection};
adaptive novelty detection with FDR guarantees appears in \citet{aos2024adaptive}, and
conformal selection under hierarchical structure in \citet{leeren2025}. We use this
machinery as given. Our per-decision FPR statement
(Proposition~\ref{prop:validity}) is a direct consequence of exchangeability and is not
claimed as a contribution. What is new is (i) the analysis of certified admission
inside the poisoning dynamics: per-batch FDR control alone provably does
not prevent long-run poisoning (Remark~\ref{rem:fdr-not-enough}); what does is
severing the evidence loop, and we prove exactly that (Lemma~\ref{lem:ebh},
Theorem~\ref{thm:gating}); and (ii) the CDC construction, which is a calibration
procedure, not a selection procedure.

\paragraph{Contaminated calibration.}
\citet{bashari2025} study conformal outlier detection with a contaminated
reference set, obtaining conservative validity and using labelled outliers for
power. CDC differs in all three coordinates that matter at deployment: it is label-free;
it treats contamination of the stream used for recalibration under simultaneous
ID drift (their reference set is fixed); and it comes with an explicit power bound plus
a matching impossibility result. Concurrent work \citep{wang2026trimming} analyzes when
trimming a contaminated calibration set preserves coverage (a retained-law
diagnostic); CDC deliberately takes the opposite route, no sample selection at all,
correcting the quantile level instead, which is what allows a guarantee under
drift, where trimming-style cleaning provably recovers little
(Section~\ref{sec:setting}). Storey-type null-proportion estimation
\citep{storey2002} and DKW-based quantile certificates are classical; the observation
that calibration drift biases the Storey statistic in the conservative direction
(Theorem~\ref{thm:storey}) appears to be new. Recalibrating a conformal cutoff from
unlabelled test data under pure distribution shift is studied by
\citet{yilmaz2022qtc}. A contaminated stream defeats shift-only recalibration, because
drift and contamination are confounded in unlabelled data, which is exactly the content
of Theorem~\ref{thm:impossible}.

\paragraph{Impossibility results.} \citet{fang2022learnable} prove that OOD detection
is not learnable without restrictions on the out-distribution.
Theorem~\ref{thm:impossible} is complementary. It concerns label-free calibration
rather than learnability, and its construction follows the
least-favorable-contamination tradition of \citet{huber1965} and the
tail-irreducibility phenomenon of semi-supervised novelty detection
\citep{blanchard2010}. The closed-form ceiling $\kappa$ and its empirical match by a
deployed procedure appear to be new.

\paragraph{Urn processes and stochastic approximation.}
Our dynamical analysis uses generalized P\'olya urns via the ODE method
\citep{renlund2010, benaim1999, pemantle2007, laruelle2013}. These tools are standard
in probability, and phase transitions in reinforced urns are themselves known:
\citet{laruelle2019nonlinear} exhibit the passage from a single attracting equilibrium
to a two-attractor system in nonlinear randomized urns. The novelty here is not the
existence of urn phase transitions but their identification inside a deployed detector
class, where the admission kernel is measurable, the measured coefficients place a
deployment on the phase diagram, and a certified gate removes the transition.

\paragraph{Relation to our own prior submission.} A companion manuscript (under review)
studies the static, batch-level transductive question: how much OOD information
the spectrum of a single unlabelled test batch carries, analyzed with spiked
random-matrix theory. It involves no adaptation over time, no memory banks, and no
calibration-under-drift claims; conversely the present paper uses no spectral or
random-matrix machinery. The two share no theorems and no experimental protocol (batch
scoring there; contaminated streaming with closed-loop state here). Where that paper
uses an AdaODD-style method as a comparison baseline, here the adaptive detectors
are the object of study, their closed-loop admission dynamics and failure
modes. Either paper stands independently of the other.

\section{Setting and the self-poisoning phenomenon}\label{sec:setting}

\paragraph{Protocol.}
A detector observes batches $X_1, X_2, \dots$ of feature vectors from a stream in which
each point is ID with probability $1-\pi$ and OOD with probability $\pi$, arriving
either i.i.d.\ or in bursts (contiguous OOD blocks, matching how outliers arrive in
production). The detector outputs per-point scores (larger = more OOD) and binary flags
targeting FPR $\le\alpha$; it may adapt internal state but never sees labels. We
evaluate realized FPR, TPR, AUROC, and, for adaptive banks, the impurity $\rho_t$
of the bank. Full experimental detail in Section~\ref{sec:experiments}. In brief, 96
settings over four encoder families (trained vision, frozen foundation, text, document),
$\pi\in\{0.01,0.05,0.1,0.5\}$, both orderings, multiple seeds, $\alpha=\delta=0.10$.

\begin{figure}[t]
\centering
\includegraphics[width=\textwidth]{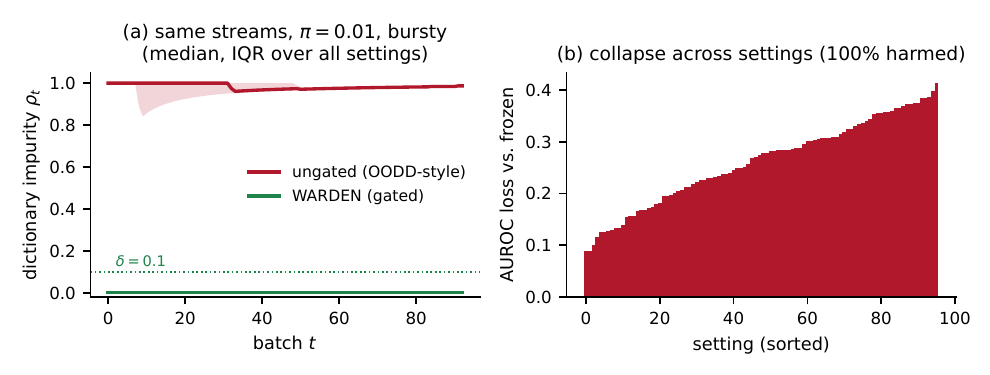}
\caption{Self-poisoning is universal, not a corner case. (a)~Dictionary
impurity $\rho_t$ on identical bursty streams at $\pi=0.01$ (median and interquartile
range over all 96 settings $\times$ 5 seeds): the ungated detector's ``OOD'' dictionary
is $\ge90\%$ ID within a few batches, while WARDEN's certified admission keeps impurity
below its budget $\delta=0.1$ throughout. (b)~Resulting AUROC loss of the ungated
detector relative to its own frozen baseline, one bar per setting (sorted): all
96 settings are harmed at $\pi=0.01$ bursty, by up to $0.41$ AUROC.}
\label{fig:phenomenon}
\end{figure}

\paragraph{Phenomenon 1: ungated adaptation collapses.}
In bursty streams with $\pi\le 0.1$, the ungated OOD-dictionary detector loses
$0.163$ mean AUROC relative to its own frozen baseline (up to $0.28$ on individual
families; below chance on near-OOD pairs), and its dictionary impurity exceeds $0.9$.
The ``OOD'' dictionary is more than $90\%$ ID. The ungated ID-bank detector loses
$0.026$ AUROC through the mirror-image mechanism. On a full replication over all 96
settings using five fresh prime seeds never touched during development, the
collapse reproduces (\CONFcollapseOODD{} and \CONFcollapseADA{} respectively).
Figure~\ref{fig:phenomenon} shows the phenomenon. On identical streams, the ungated
dictionary saturates with ID junk within a few batches while the gated detector of
Section~\ref{sec:gating} stays below its budget, and at $\pi=0.01$ bursty every
one of the 96 settings is harmed.

\paragraph{Phenomenon 2: static calibration is stale.}
A detector calibrated once on train-ID data, the only data a real deployment has, realizes FPR $=0.194$ at $\alpha=0.10$ ($1.94\times$ nominal; range $1.3$--$2.2\times$
across families), against $0.100$ for an oracle calibrated on held-out test-ID. The
drift is a per-dimension variance change in whitened space. Re-centering the stale
reserve recovers nothing, an oracle affine (mean+variance) correction recovers
$\sim\!75\%$ of the gap, and an online-estimated affine correction only
$\sim\!38\%$, motivating a calibration procedure that does not require estimating the
drifted density at all (Section~\ref{sec:cdc}).

Phenomena 1 and 2 bracket the design space: adapt ungated and collapse, or freeze and
drift out of calibration. The remainder of the paper shows both failures are provable,
both fixes are certifiable, and the residual power cost is information-theoretically
necessary.

\paragraph{Scope.} Everything that follows is stated at the level of feature vectors
and admission dynamics. Nothing in the theory references pixels, tokens, or page
layouts. Its premises are structural. A detector self-updates from its own unlabelled
decisions, the stream mixes two populations, and the admission kernel depends on the
bank state. Any deployment with these ingredients (transaction fraud monitoring,
intrusion detection, and clinical anomaly screening are natural instances) is in scope
in principle. Our evidence covers four encoder families over vision, text, and document
data. Carrying the guarantees to a new domain requires re-measuring the kernel of
Section~\ref{sec:theory} on that domain's features, not new theory.

\section{A sharp-threshold theory of self-poisoning}\label{sec:theory}

\subsection{The admission process}

Let $D_t$ denote the adaptive bank after batch $t$, containing $n_t$ points of which
$m_t$ were wrongly admitted (for an OOD dictionary, wrong = ID); write
$\rho_t = m_t/n_t$. At batch $t$ the detector admits $a_t \ge 0$ points, $w_t$ of them
wrong. All quantities are adapted to the filtration $\mathcal F_t$ generated by the
stream and the detector's state.

\begin{assumption}[Kernel]\label{ass:kernel}
The admission size $a_t$ is $\mathcal F_{t-1}$-measurable (predictable), and there is a
Lipschitz $q:[0,1]\to[0,1]$ such that the false-admission proportion obeys
$\mathbb E\bigl[w_t/a_t \mid \mathcal F_{t-1}\bigr] = q(\rho_{t-1})$ on $\{a_t>0\}$;
that is, $q(\rho)$ is the expected fraction of a batch's admissions that are wrong,
given current impurity $\rho$.
\end{assumption}

Predictability of $a_t$ holds by construction for the detectors under study:
AdaODD- and OODD-style rules admit a fixed fraction of each batch, so $a_t \equiv
\lceil q_{\mathrm{adm}} K\rceil$ is a constant. (For rules with random data-dependent
$a_t$ the mean field is instead driven by the admission-weighted kernel $\bar q(\rho) =
\mathbb E[w_t\mid\mathcal F_{t-1}]/\mathbb E[a_t\mid\mathcal F_{t-1}]$; all results below
hold verbatim with $q$ read as $\bar q$, and our pooled empirical fits estimate exactly
this weighted kernel.) The single-argument form $q(\rho)$ is an idealization for bursty
streams, where the instantaneous proportion is modulated by the burst phase; there
$q(\rho)$ is the phase-averaged kernel, the ODE method applies in its Markov-modulated
form \citep[Ch.~8]{benaim1999}, and our fits (Section~\ref{sec:experiments}) estimate
precisely this average. Assumption~\ref{ass:kernel} is an empirical claim we test, not a
convenience: on all 96 settings the measured kernel is affine in $\rho$ with $R^2 \ge
0.996$. The mechanism behind the $\rho$-dependence is the contrast score: wrong points
in the dictionary raise the OOD-similarity of nearby ID points, so admission mistakes
reinforce.

\begin{assumption}[Admission rate and noise]\label{ass:rate}
$1 \le a_t \le a_{\max}$ on the event $a_t > 0$, admissions occur in a positive fraction
of batches, $q$ is differentiable in a neighborhood of each unstable zero of
$h(\rho)=q(\rho)-\rho$, and $\mathrm{Var}(w_t/a_t \mid \mathcal F_{t-1}) \ge \sigma_0^2
> 0$ on $\{a_t>0\}$ in a neighborhood of every such zero. (On $\{a_t=0\}$ we set
$\xi_t := 0$; the recursion does not move.)
\end{assumption}

While the bank is growing ($n_t\uparrow$), a one-line computation from
$\rho_t = \tfrac{m_{t-1}+w_t}{n_{t-1}+a_t}$ (Appendix~\ref{app:theory}) gives the exact
recursion
\begin{equation}\label{eq:sa}
\rho_t = \rho_{t-1} + \gamma_t\bigl[h(\rho_{t-1}) + \xi_t\bigr],
\qquad h(\rho) := q(\rho) - \rho,
\end{equation}
with step $\gamma_t = a_t/n_t$ and noise $\xi_t = w_t/a_t - q(\rho_{t-1})$, $|\xi_t|\le
1$. Because $a_t$ is predictable, $n_t = n_{t-1}+a_t$ and hence $\gamma_t$ are
$\mathcal F_{t-1}$-measurable, so $\gamma_t\xi_t$ is a genuine martingale-difference
perturbation: a generalized P\'olya urn in stochastic-approximation form. The state is a
proportion, so $q,\rho\in[0,1]$ and the drift is never vacuous.

\subsection{Convergence, threshold, and bifurcation}

\begin{theorem}[Almost-sure convergence to mean-field equilibria]\label{thm:converge}
Under Assumptions~\ref{ass:kernel}--\ref{ass:rate}, on the event that admissions
accumulate ($n_t\to\infty$), $\rho_t$ converges almost surely to the zero set of $h$;
if the zeros are isolated, $\rho_t\to Z$ where $Z$ is a single zero, and
$\Pr(Z = \rho^u) = 0$ for every linearly unstable zero $\rho^u$ (i.e.\ with
$q'(\rho^u) > 1$) at which the noise floor of Assumption~\ref{ass:rate} is active.
\end{theorem}

\begin{corollary}[Reproduction number for affine kernels]\label{cor:affine}
Let $q(\rho) = \min\{a + b\rho,\, 1\}$ with $a\in(0,1)$, $b\ge 0$, and define
$R_0 := b$, and work on the admission-accumulation event of
Theorem~\ref{thm:converge}. If $R_0 < 1$, then $\rho_t \to \rho^* = a/(1-b)\wedge 1$
a.s.\, partial poisoning with amplification factor $(1-b)^{-1}$ when $a < 1-b$,
complete poisoning when $a \ge 1-b$. If $R_0 \ge 1$, then $\rho_t \to 1$ a.s.
(complete poisoning), for every $a>0$ however small.
\end{corollary}

\begin{figure}[t]
\centering
\includegraphics[width=\textwidth]{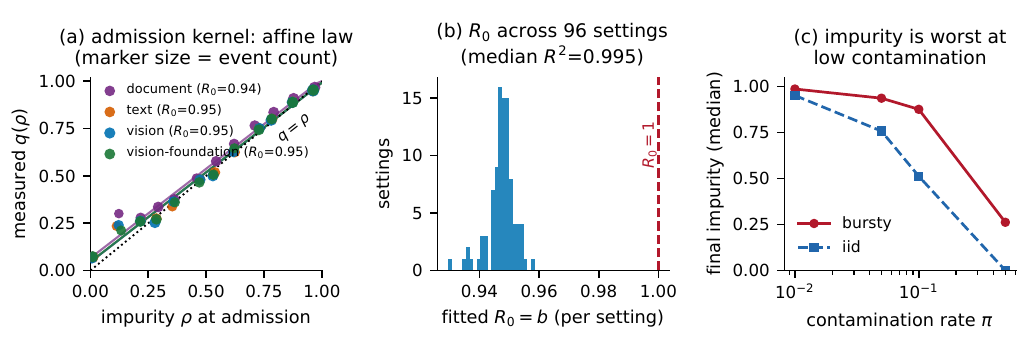}
\caption{The admission kernel is an affine law, measured, not assumed.
(a)~False-admission proportion vs.\ instantaneous impurity $\rho$, pooled over $4.3
\times 10^6$ admission events and binned (12 bins, count-weighted; marker size $\propto$
log event count), with weighted affine fits per encoder family: aggregate slope $R_0 = 0.94$--$0.95$
everywhere, just below the critical diagonal $q=\rho$ (a protocol signature rather
than an encoder constant). The fits are
admission-weighted over the deployment grid; $\pi$-conditional coefficients are in
Appendix~\ref{app:exp}.
(b)~Per-setting fits: $R_0$ concentrates in $[0.929, 0.959]$ with median $R^2=0.995$
across all 96 settings, the affine form of Assumption~\ref{ass:kernel} is not a
convenience. (c)~Median final impurity of the ungated dictionary vs.\ $\pi$: because
$a(\pi)$ grows as streams become more ID-dominated, impurity is highest at the
smallest contamination rates, and bursty ordering saturates the dictionary already at
$\pi=0.01$.}
\label{fig:kernel}
\end{figure}

Empirically the collapse operates through the first branch at its boundary: every
measured setting has $R_0 \in [0.929, 0.959] < 1$ but $a \ge 1-b$, so
$\rho^* = 1$, the supercritical branch $R_0>1$ is a theoretical completion our
settings do not reach. Figure~\ref{fig:kernel} shows the measured law: binned
false-admission proportions hug the affine fits with $R_0\approx0.94$--$0.95$ in every
family (panel a), the per-setting $R_0$ distribution concentrates tightly below the
critical value (panel b), and the resulting impurity is worst at the lowest
contamination rates (panel c), the counterintuitive inversion that makes bursty
low-$\pi$ deployment, the realistic regime, the dangerous one.

The contamination rate enters through the batch composition: $a=a(\pi)$, $b=b(\pi)$.
The operational critical rate, the quantity we measure, is
$\pi_c := \inf\{\pi : \rho^*(\pi) \ge 1/2\}$, which under Corollary~\ref{cor:affine} is
determined by the measured kernel coefficients. In bursty streams the pure-ID batches
make $a(\pi)$ large already at small $\pi$, which is why the measured knee sits at
$\pi_c \approx 0.01$: burstiness, not high contamination, is the dangerous regime.
Empirically the affine form of Assumption~\ref{ass:kernel} is not merely convenient.
The measured kernel fits with $R^2 \ge 0.996$ across all four encoder families with
slope $R_0 \approx 0.947$ just below critical (Section~\ref{sec:experiments}), and
$\rho^* = a/(1-b)$ sits at the saturation boundary, so ungated adaptation is
essentially always in the complete-poisoning regime once bursts appear. The
concentration of the aggregate slope near one across families is itself informative.
It is a signature of the fixed-fraction, bank-relative admission protocol, which pins
the composition of admitted points close to the bank's current composition, not an
encoder-level constant. Conditioning on $\pi$ decomposes the same law into the pair
$(a(\pi), b(\pi))$ that the theory actually consumes, intercept-driven at low
contamination and slope-driven at high (Appendix~\ref{app:exp}), and the severed
design of Section~\ref{sec:gating} exhibits no measurable $\rho$-dependence at all.
The detector class is near-critical by design, and the contamination rate sets which
coordinate tips it.

\begin{proposition}[Fold bifurcation and hysteresis for saturating kernels]
\label{prop:fold}
Consider the mean-field flow $\dot\rho = h(\rho;\theta) = q(\rho;\theta)-\rho$ for a
kernel family $q(\rho;\theta)$ that is: (i) $C^2$ in $\rho$ and $C^1$ in $\theta$, with
$0<q(0;\theta)$ and $q(1;\theta)<1$; (ii) strictly increasing in $\rho$ with
$\partial_\rho q$ strictly unimodal (a single inflection, ``sigmoidal''); (iii)
strictly increasing in $\theta$ with $\partial_\theta q > 0$; (iv) steep enough somewhere:
$\sup_\rho \partial_\rho q(\rho;\theta^*) > 1$ at some $\theta^*$, with a unique low
(resp.\ high) diagonal crossing at the sweep's lower (resp.\ upper) end; and (v)
non-degenerate at tangencies ($\partial^2_{\rho\rho}q \neq 0$ whenever
$q=\rho,\ \partial_\rho q =1$). Then there exist $\theta_- < \theta_+$ such that the
equilibrium set passes from a unique low equilibrium ($\theta<\theta_-$), through a
saddle-node pair creating bistability ($\theta_-<\theta<\theta_+$), to a unique high
equilibrium ($\theta>\theta_+$); under a quasi-static $\theta$-sweep the attained
equilibrium of the flow jumps discontinuously at $\theta_+$ (upward) and $\theta_-$
(downward): hysteresis.
\end{proposition}

Unimodality of $\partial_\rho q$ bounds the diagonal crossings by three (the sign
pattern of $h'$ has at most two changes), which is what rules regimes beyond the
fold sequence out. The proposition concerns the deterministic mean field; for the
stochastic process, Theorem~\ref{thm:converge} pins $\rho_t$ to the corresponding stable
branch between fold points on the decreasing-step time scale.

\begin{proposition}[Capped-bank regime, uniform eviction]\label{prop:fifo}
Once the bank reaches its eviction cap $N$, suppose admissions are constant
($a_t \equiv \bar a$, step $\gamma = \bar a/N$; true for the fixed-fraction rules
studied here), eviction removes a uniformly random subset of $\bar a$ incumbents per
admission, and the affine kernel is unsaturated ($a+b\le 1$; in the saturated case
$\rho_t\to1$ by Corollary~\ref{cor:affine} and there is nothing to prove). Then for
$b<1$:
$\;\bigl|\mathbb E[\rho_t] - \rho^*\bigr| \le (1-\gamma(1-b))^{t-t_0}
|\rho_{t_0}-\rho^*|$ and
$\limsup_t \mathrm{Var}(\rho_t) \le \gamma\,\sigma_{\max}^2/(2(1-b) - \gamma(1-b)^2)$,
where $\sigma^2_{\max} := \sup \mathrm{Var}\bigl(w_t/a_t - \rho^{\mathrm{ev}}_t \mid
\mathcal F_{t-1}\bigr) \le 1$: the process forgets its past geometrically and fluctuates
within $O(\sqrt{\gamma})$ of the equilibrium.
\end{proposition}

Uniform eviction makes the evicted block's conditional mean impurity exactly
$\rho_{t-1}$, which is what the contraction uses. Literal FIFO (our implementation)
instead evicts the oldest block, whose impurity is a lagged functional of the
window; the recursion acquires a delay term with the same unique fixed point and an
$O(\gamma)$ bias floor, and the two rules are empirically indistinguishable in all our
measurements. We state the clean result for uniform eviction and treat FIFO as its
delay-perturbed variant.

\subsection{Gated admission removes the transition}\label{sec:gating}

The detector's decisions themselves can be certified by conformal validity: this part is
classical and stated only for completeness.

\begin{proposition}[Per-decision FPR validity; folklore]\label{prop:validity}
Let $R$ be a reserve of ID points exchangeable with a true-ID test point $x$, and let
the score function used at step $t$ be $\mathcal F_{t-1}$-measurable. Then the one-sided
conformal p-value of $x$ against $R$ is super-uniform, so flagging at $p\le\alpha$
realizes $\Pr(\mathrm{flag}\ x) \le \alpha$ for every $t$ and every adaptation rule.
\end{proposition}

The substantive question is the admission side. The collapse mechanism of
Theorem~\ref{thm:converge} is the $\rho$-dependence of the kernel: admission evidence
computed against the mutable bank makes mistakes self-reinforcing. WARDEN's design
principle is to sever this loop (Figure~\ref{fig:arch}b). Admission evidence is
computed only
against the frozen reserve with the frozen base scorer, the dictionary is never
consulted for admission, and the e-BH gate then adapts how many points that evidence
admits. Two guarantees follow. First, the classical one: e-BH at level $\delta$ controls
the per-batch expected false-admission proportion,
$\mathbb E[w_t/\max(a_t,1)\mid\mathcal F_{t-1}]\le\delta$, under arbitrary within-batch
dependence \citep{wang2022ebh}. We note honestly that this per-batch FDR bound
alone does not cap the long-run impurity ratio: adversarially structured
e-values can satisfy it while admitting only wrong points, rarely. What removes the
bifurcation is the severed feedback, quantified next.

\begin{lemma}[$\rho$-independent wrong-admission intensity]\label{lem:ebh}
Let each batch of size $K$ be gated by e-BH at level $\delta$ over $e_i =
a\,p_i^{a-1}$, $a\in(0,1)$, where $p_i$ is the conformal p-value of point $i$ against a
frozen ID reserve of size $m$ under a frozen scorer, and let the stream's true-ID points
be exchangeable with the reserve. Every e-BH admission requires $e_i \ge K/(\delta a_t)
\ge 1/\delta$, i.e.\ $p_i \le \kappa(\delta) := (a\delta)^{1/(1-a)}$. Consequently, with
probability $\ge 1-\eta$ over the reserve draw, for every batch $t$,
\[
\mathbb E[w_t \mid \mathcal F_{t-1}] \;\le\; K\,\bar\kappa(m,\kappa,\eta)
\;=:\; \bar W,
\qquad
\bar\kappa := \max\bigl\{u :
\Pr\bigl[\mathrm{Bin}(m,u) < \lceil\kappa(m+1)\rceil\bigr] \ge \eta\bigr\},
\]
the exact binomial upper confidence limit for the reserve's $\kappa$-tail quantile
(asymptotically $\bar\kappa = \kappa + O(\sqrt{\kappa\log(1/\eta)/m})$), uniformly in
the dictionary state $\rho$, the contamination rate $\pi$, the stream ordering, and the
within-batch dependence.
\end{lemma}

\begin{theorem}[Severed feedback: no supercritical branch]\label{thm:gating}
Under Lemma~\ref{lem:ebh}, on the reserve event: (i) wrong admissions accrue at a
$\rho$-independent rate, $\limsup_t m_t/t \le \bar W$ a.s.; (ii) on streams where total
admissions accrue at rate $\liminf_t n_t/t \ge r > 0$ (e.g.\ any stream whose OOD mass
the gate detects at positive rate), $\limsup_t \rho_t \le \bar W/r$ a.s.; and (iii) the
gated kernel obeys $\bar q_{\mathrm{gated}}(\rho) \le \bar W/(\bar W + r)$ for all
$\rho$, it does not increase with impurity, so the reinforcement mechanism behind
Corollary~\ref{cor:affine} and Proposition~\ref{prop:fold} is structurally absent and no
supercritical branch exists, for every $\pi$ and every ordering.
\end{theorem}

\begin{corollary}[Adversarial contamination]\label{cor:adversarial}
Let the reserve be collected before deployment. Suppose the contaminated points of
each batch are chosen by an adversary that observes the full history, the reserve, and
the detector's code, and that picks the number, the placement, and the feature vectors
of those points arbitrarily, subject only to the true-ID points remaining exchangeable
with the reserve. Then Proposition~\ref{prop:validity}, Lemma~\ref{lem:ebh}, and
Theorem~\ref{thm:gating} hold verbatim, with the same constants.
\end{corollary}

The reason (Appendix~\ref{app:theory}) is that the guarantees rest on two ingredients
only, the exchangeability of true-ID points with the frozen reserve under a predictable
scorer, and the absolute admission threshold $e\ge1/\delta$ computed on the frozen
channel. Neither ingredient depends on where the adversary puts its points, and wrong
admissions are by definition true-ID admissions, whose evidence law the adversary does
not control. The corollary does not say attacks are harmless. An adversary can suppress
the dictionary channel's contribution to decisions, a power effect rather than a
validity effect, and an empirically negligible one (Appendix~\ref{app:c2}).
Appendix~\ref{app:attack} stress-tests the corollary directly: under a gray-box
tail-mimicry attack of increasing strength, realized FPR and dictionary impurity are
unchanged, and the only quantity the adversary moves is detection power, which decays
to the confusable floor of Theorem~\ref{thm:impossible}. What it
cannot do is start the cascade or break a certificate.

At our operating point ($a=0.1$, $\delta=0.1$, $K=64$, $m=1500$, $\eta=0.05$):
$\kappa \approx 0.006$, exact binomial limit $\bar\kappa \approx 0.0096$, and $\bar W
\approx 0.62$ wrong admissions per batch, an
a-priori ceiling; the measured impurity is far below it (max $0.111$ over all $6{,}960$
cells, against ungated impurity $>0.9$), and measured per-batch FDR respects $\delta$.
The lemma's exchangeability premise is met in our protocol because the admission reserve
is held-out eval-ID; with a stale reserve under train$\to$test drift the
intensity inflates by exactly the drift deficiency of Section~\ref{sec:cdc}, which is
where CDC takes over, the two components compose. Theorem~\ref{thm:gating} is a
statement about the closed-loop dynamics: certified admission does not merely
make fewer mistakes, it deletes the mechanism by which mistakes compound, because the
admission channel never reads the object it feeds. Empirically the gated detector's
AUROC never departed from the frozen baseline, across all cells including every
collapsing one.

\section{CDC: certified decontaminated calibration}\label{sec:cdc}

We now turn to Phenomenon 2. The deployment has: a stale reserve
$R_{\mathrm{tr}}$ of train-ID scores; an unlabelled stream window
$Z = (s_1,\dots,s_n)$ of scores drawn from the mixture
$\Fmix = (1-\pi)\Fid + \pi\Food$, where $\Fid$ is the current (drifted) test-ID
score law; and no labels, ever. The goal is a threshold $\hat\tau$ with
$\Pid(s > \hat\tau) \le \alpha$.

\paragraph{Algorithm (CDC).} With window size $n$, Storey parameter $\lambda$
(we use $\lambda=1/2$; for even reserve sizes this sits between adjacent grid points
of Theorem~\ref{thm:storey}, shifting $\hat\pi_\lambda$ by $O(1/m)$, which the
Hoeffding slack absorbs),
confidence $\eta$:
\begin{enumerate}[leftmargin=1.6em,itemsep=0pt,topsep=2pt]
\item Compute conformal p-values $p_i$ of the window scores against the stale reserve
$R_{\mathrm{tr}}$, and the Storey statistic
$\hat\pi_\lambda = 1 - \frac{\#\{p_i > \lambda\}}{n(1-\lambda)}$; set
$\hat\pi_{\mathrm{up}} = \bigl[\hat\pi_\lambda +
\tfrac{1}{1-\lambda}\sqrt{\tfrac{\log(2/\eta)}{2n}}\bigr]_0^{1}$.
\item Output $\hat\tau = \hat F^{-1}_{\mathrm{mix}}
\bigl(1 - \alpha(1-\hat\pi_{\mathrm{up}}) + \eps_n\bigr)$, where $\hat F_{\mathrm{mix}}$
is the empirical CDF of the window and $\eps_n = \sqrt{\log(2/\eta)/(2n)} + 1/n$ (the
$1/n$ absorbs empirical-quantile discreteness).
\item Refresh the window and repeat; the threshold applied to batch $t$ is computed from
batches $<t$ (decide-then-update).
\end{enumerate}

Note what CDC does not do: it never tries to identify which points are ID, never
estimates the drifted density, and never touches the bank machinery of
Section~\ref{sec:theory}, it corrects the quantile level instead of the
sample. This is why its guarantee survives exactly where sample-cleaning heuristics
(Section~\ref{sec:setting}) fail.

\begin{theorem}[Validity]\label{thm:cdc}
Let $\pi<1$, let the window be an i.i.d.\ draw from $\Fmix$, and let
$\hat\pi_{\mathrm{up}} \ge \pi$ hold on an event of probability $\ge 1-\eta_1$. Then
with probability at least $1-\eta/2-\eta_1$,
\[
\Pid(s > \hat\tau)\;\le\; \frac{\Pmix(s>\hat\tau)}{1-\pi}
\;\le\; \frac{\alpha(1-\hat\pi_{\mathrm{up}})}{1-\pi} \;\le\; \alpha .
\]
No assumption on $\Food$ is required for validity. If the requested level
$1-\alpha(1-\hat\pi_{\mathrm{up}})+\eps_n$ exceeds $1$ (possible when
$\hat\pi_{\mathrm{up}}$ is clipped near $1$), set $\hat\tau=+\infty$ (flag nothing):
trivially valid, with power governed by Theorem~\ref{thm:power}'s degenerate case.
\end{theorem}

\begin{theorem}[Drift-conservativity of the Storey estimate]\label{thm:storey}
Let the p-values be conformal against the realized stale reserve $R_{\mathrm{tr}}$ of
size $m$, with continuous scores and $\lambda$ on the reserve grid
$\{k/(m+1)\}$. Conditionally on $R_{\mathrm{tr}}$, define the (reserve-conditional)
drift deficiency $\eps_{\mathrm{dr}}(R) := (1-\lambda) - \Pid(p > \lambda \mid R)$ and
leakage $\kappa_\lambda(R) := \Pood(p>\lambda \mid R)$.
(a) Conditional conservativity. If
\[
(1-\pi)\,\eps_{\mathrm{dr}}(R) \;\ge\; \pi\, \kappa_\lambda(R)
\tag{A2}
\]
then $\mathbb E[\hat\pi_\lambda \mid R] \ge \pi$, and with the Hoeffding slack of Step 1
(valid conditionally on $R$, since the window is i.i.d.\ given $R$),
$\Pr(\hat\pi_{\mathrm{up}} < \pi \mid R) \le \eta/2$. (A2) holds whenever
$\pi \le \pi_\dagger(R) := \eps_{\mathrm{dr}}(R) / (\eps_{\mathrm{dr}}(R) +
\kappa_\lambda(R))$, with the convention $\pi_\dagger := 1$ when both quantities vanish
(far OOD with no drift: (A2) is then $0\ge0$, true for all $\pi$).
(b) Dominance makes (A2) population-checkable. If test-ID scores stochastically
dominate train-ID scores (the measured drift direction), then the reserve-averaged
deficiency satisfies $\mathbb E_R[\eps_{\mathrm{dr}}(R)] \ge 0$ (grid $\lambda$ makes
the exchangeable baseline exact), and by DKW on the reserve, with probability
$\ge 1-2\eta_R$ over the reserve draw,
$\eps_{\mathrm{dr}}(R) \ge \bar\eps_{\mathrm{dr}} - \eps_m$ and $\kappa_\lambda(R) \le
\bar\kappa_\lambda + \eps_m$, $\eps_m = \sqrt{\log(2/\eta_R)/(2m)}$, where bars denote
reserve-averaged quantities. Hence the population condition
$(1-\pi)(\bar\eps_{\mathrm{dr}} - \eps_m) \ge \pi(\bar\kappa_\lambda + \eps_m)$ implies
(A2) with probability $\ge 1-2\eta_R$, and then
$\Pr(\hat\pi_{\mathrm{up}} < \pi) \le \eta/2 + 2\eta_R$.
\end{theorem}

Condition (A2) is the honest boundary of the method: the drift-induced sub-uniformity
must dominate the OOD leakage above $\lambda$, and it holds automatically in the field's
dominant regime (far OOD: $\kappa_\lambda \approx 0$, (A2) free). Because drift and
contamination are indistinguishable from unlabelled data (Theorem~\ref{thm:impossible}),
(A2) cannot be verified label-free. Some separation assumption is logically required
of any label-free method, and (A2) is a mild, interpretable instance. We therefore
state the guarantee in two tiers and regard the first as primary. The unconditional
tier assumes nothing about the outliers: set $\hat\pi_{\mathrm{up}}=\pi_{\max}$ from a
governance policy (``we design for at most $\pi_{\max}$ contamination''), and
Theorem~\ref{thm:cdc} holds for all $\pi\le\pi_{\max}$, trading a known amount of
power for an assumption-free certificate. The adaptive tier uses the Storey estimate
and is certified under (A2). It is the variant we evaluate, and its empirical record
(certification on every tested drift-affected cell, including the bursty cells that
violate the i.i.d.\ window assumption, Section~\ref{sec:experiments}) is evidence
about (A2) in practice, not a substitute for the assumption.

\begin{theorem}[Power]\label{thm:power}
Let $\tau^* = \Fid^{-1}(1-\alpha)$ be the oracle threshold and suppose $\Fmix$ has
density bounded below by $\kappa>0$ on the deterministic interval $[\tau^*,
\Fmix^{-1}(1-\alpha(1-\hat\pi_{\mathrm{up}})+2\eps_n)]$ (which contains $\hat\tau$ on
the DKW event; if $\hat\tau < \tau^*$ the TPR gap is nonpositive and the bound is
trivial). Then with probability at least $1-\eta-\eta_1$,
\[
\mathrm{TPR}(\tau^*) - \mathrm{TPR}(\hat\tau) \;\le\;
\frac{\bar f_{\mathrm{OOD}}}{\kappa}\,
\Bigl( \alpha\,\hat\pi_{\mathrm{up}} + (1-\alpha)\,\pi + 2\eps_n \Bigr),
\]
where $\bar f_{\mathrm{OOD}}$ bounds the OOD score density on the same interval. In the
absence of material drift ($\bar\eps_{\mathrm{dr}}\to0$, so
$\hat\pi_{\mathrm{up}} = O(\pi + n^{-1/2})$) the power loss vanishes as $\pi \to 0$ at
rate $O(\pi + n^{-1/2})$; under drift, $\hat\pi_{\mathrm{up}}$ carries the systematic
term $\bar\eps_{\mathrm{dr}}/(1-\lambda)$ and the loss has a non-vanishing floor
$\approx \alpha\,\bar\eps_{\mathrm{dr}}/(1-\lambda) \cdot \bar
f_{\mathrm{OOD}}/\kappa$, the certified price of drift-robustness, in line with
Theorem~\ref{thm:impossible}.
\end{theorem}

\begin{proposition}[Block-dependent windows]\label{prop:block}
Suppose the window is a concatenation of independent blocks of lengths $l_j \le
L$ (each block i.i.d.\ internally or not, only cross-block independence is used; the
natural model for bursty arrival, where each burst is one block). Let
$n_{\mathrm{eff}} := n^2/\sum_j l_j^2 \ge n/L$. Then Theorems~\ref{thm:cdc}
and~\ref{thm:storey} hold with the window deviation terms replaced by
\[
\eps(n_{\mathrm{eff}}) \;=\;
\sqrt{\frac{\log\bigl(2(k{+}1)/\eta\bigr)}{2\,n_{\mathrm{eff}}}} + \frac1k
\quad\text{for any grid size } k \text{ (take } k=\lceil\sqrt{n_{\mathrm{eff}}}\,\rceil),
\]
i.e.\ deviations at the effective sample size $n_{\mathrm{eff}}$ up to a logarithmic
factor. A deployment that inflates its slack to $\eps(n_{\mathrm{eff}})$ retains the
certificate under $L$-block dependence.
\end{proposition}

The proof (Appendix~\ref{app:cdc}) uses a quantile grid with per-point Hoeffding bounds
over the independent blocks (with the length-weighted bounded-difference constants
$l_j/n$), plus monotonicity, not a naked McDiarmid, whose centering term is not
negligible at this scale; exchangeability across blocks would not suffice (a de
Finetti mixture defeats any such concentration). This worst-case correction is
deliberately loose: at $n=4000$, $L=100$ it inflates the slack to $\approx0.4$, far
larger than the empirical margin. Yet the i.i.d.-calibrated CDC (no inflation) already certifies $100\%$
of bursty cells at FPR $0.041$, the realized bursts are far from the adversarial
block arrangement the bound guards against, so the correction, while available for a
worst-case guarantee, is not needed in practice. Closing this gap with a tight
dependent-window certificate (e.g.\ via block-bootstrap quantiles) is the paper's main
theoretical loose end, stated as such.

\begin{figure}[t]
\centering
\includegraphics[width=\textwidth]{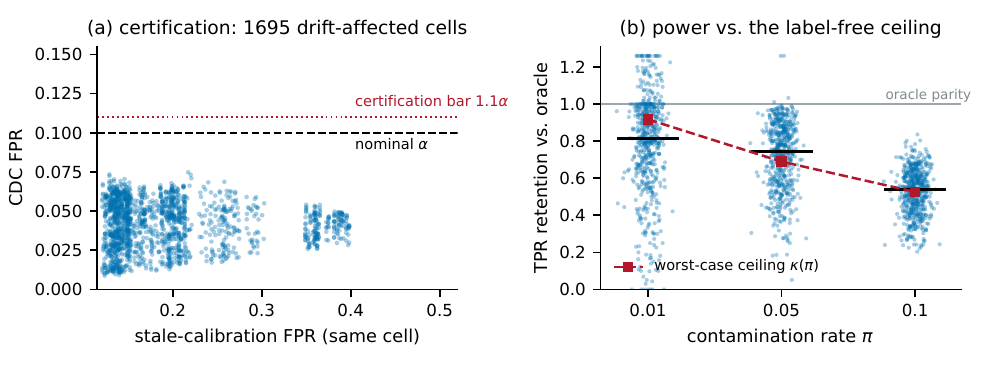}
\caption{CDC certifies every drift-affected cell and pays a bounded, predicted
power price. (a)~Per-cell realized FPR of CDC against the
stale calibration on the same cell, for all $1{,}695$ drift-affected cells: the
stale detector runs at up to $4\times$ nominal while every CDC cell sits below the
certification bar $1.1\alpha$ (max $0.075$). (b)~Per-cell TPR retention vs.\ the
oracle by contamination rate, with the worst-case label-free ceiling $\kappa(\pi) =
\alpha/(\pi+\alpha(1-\pi))$ of Theorem~\ref{thm:impossible} (red). Medians (black
bars: $0.81$, $0.75$, $0.54$) track the ceiling's shape; individual benign cells may
exceed $\kappa$ and even parity with the oracle ($14\%$ of $\pi=0.01$ cells, where
small oracle TPRs make the ratio noisy; display clipped at $1.26$), the ceiling
binds guarantees on adversarial instances, not benign realizations
(Remark~\ref{rem:optimal}).}
\label{fig:cdc}
\end{figure}

\paragraph{Measured behavior.} Across all 96 settings, on the $1{,}695$ cells
where drift is material (stale FPR $\ge 1.2\alpha$) and $\pi \le 0.1$: CDC realized FPR
$\le 1.1\alpha$ in $100\%$ of cells (mean $0.042$, 95th percentile $0.063$), against
$0.194$ for the stale detector and $0.100$ for the oracle
(Figure~\ref{fig:cdc}a). The pre-registered bar $1.1\alpha$ budgets the finite-sample
slack $\eps_n$; in fact every cell landed below $\alpha$ itself (max $0.075$).

 (We report retention, the ratio to oracle TPR, for cross-setting interpretability;
Theorem~\ref{thm:power} bounds the additive gap, and the measured additive gaps are
consistent with it.) Median TPR retention against
the oracle, pooled over all drift-affected cells, was $0.665$ (foundation $0.776$, document $0.684$, vision $0.630$, text
$0.593$). On the $840$ bursty cells, violating the exchangeable-window assumption, certification was likewise $100\%$ (mean FPR $0.041$): the Storey stress test passes. A
pre-slack variant (no Hoeffding UCB) certifies $100\%$ at mean FPR $0.045$ and retains
$0.707$, quantifying the cost of the finite-sample guarantee.

\section{The price is necessary: an impossibility theorem}
\label{sec:impossibility}

\begin{theorem}[Two-world indistinguishability]\label{thm:impossible}
Fix $\alpha \in (0,1)$ and any $\pi \in (0, 1)$. Let $P$ be a continuous train-ID
score law with $\tau_B := P^{-1}(1-\alpha)$, and let $Q := \mathcal L(s \mid s >
\tau_B)$ under $P$, contamination distributed exactly like the ID tail. Set
$P' := (1-\pi)P + \pi Q$; then $P'$ stochastically dominates $P$ (a valid ``drift''),
and world $A := $ (ID $= P'$, no contamination) and world $B := $ (ID $= P$, each point
independently contaminated with probability $\pi$ by a draw from $Q$) generate
identical unlabelled stream laws.
Consequently the threshold $\hat\tau$ produced by any label-free procedure
$\mathcal A$ (a possibly randomized map from the stale reserve and the unlabelled
window to a threshold; thresholds are the natural class here since detectors flag by
score cutoff, and the indistinguishability premise applies verbatim to arbitrary
decision rules) has the same distribution in both worlds, couple the worlds
on the shared observable law, and with
$\kappa := \dfrac{\alpha}{\pi + \alpha(1-\pi)}$,
\[
\bigl\{\mathrm{FPR}_A(\hat\tau) \le \alpha\bigr\}
\;\subseteq\;
\bigl\{\mathrm{TPR}_B(\hat\tau) \le \kappa\bigr\}
\qquad\text{almost surely.}
\]
Hence every procedure with $\Pr_A[\mathrm{FPR}\le\alpha]\ge 1-\beta$ obeys
$\Pr_B[\mathrm{TPR}\le\kappa]\ge 1-\beta$: controlling FPR under drift forces
power under contamination below the ceiling $\kappa$, which satisfies $\kappa<1$ for
every $\pi\in(0,1)$ and $\kappa\downarrow\alpha$ as $\pi\uparrow 1$. A labelled oracle
achieves $\mathrm{FPR}\le\alpha$ in both worlds and $\mathrm{TPR}=1$ in world $B$ (world
$A$ has no outliers to detect).
\end{theorem}

The construction is maximally natural: from unlabelled data, ``the ID tail drifted up''
and ``outliers arrived that look like the ID tail'' are the same observation, so
no label-free rule can separate them. The two-world device itself is classical, a
least-favorable pair in the sense of \citet{huber1965} and an instance of the
irreducibility that limits semi-supervised novelty detection \citep{blanchard2010}.
What is specific to our setting is the drift-versus-contamination dichotomy and the
resulting closed form for $\kappa$.

\begin{remark}[CDC relative to the worst-case ceiling]\label{rem:optimal}
The ceiling evaluates to $\kappa=0.917$, $0.690$, $0.526$ at $\pi=0.01$, $0.05$, $0.10$
($\alpha=0.10$). CDC's measured median TPR retention over the same $\pi$-range
is $0.67$ ($0.63$ on held-out seeds), close to the mid-range ceiling $0.69$. We state
precisely what this does and does not show. It does show that the room above CDC is
bounded in the worst case: no label-free procedure that keeps the drift-validity
certificate can guarantee retention above $\kappa$, so a certified competitor can beat
CDC by at most a few points on adversarial instances. It does not show CDC is
near-optimal on our benchmarks, whose outliers are far from the adversarial ID-tail law, there the achievable retention is higher, and part of CDC's gap to the oracle is
finite-sample slack rather than information-theoretic toll. What the theorem certifies
unconditionally is the shape of the trade-off: any method retaining more power
than $\kappa$ on some instance has, on the confounded instance, given up the FPR
certificate.
\end{remark} Three consequences: (i) the stochastic
dominance/leakage condition of Theorem~\ref{thm:storey} is not an artifact, some
assumption separating the worlds is logically required by any label-free method;
(ii) CDC resolves the dilemma in the only safe direction, in world $A$ it certifies
FPR, in world $B$ it keeps FPR and pays with power on exactly the confounded band, so its measured TPR retention $<1$ is the information-theoretic toll, not an analysis
gap; (iii) methods that claim both adaptivity and full power under these conditions,
without labels or extra assumptions, cannot be correct.

\section{Experiments}\label{sec:experiments}

\paragraph{Settings.} 96 cached-feature settings spanning OpenOOD-trained
\citep{openood} vision
(ResNet-18, ViT on CIFAR-10/100 against MNIST/SVHN/DTD/TIN/Places/CIFAR cross-pairs),
frozen foundation encoders (DINOv2, CLIP), text (RoBERTa SST-2 against
20News/AGNews/MNLI), and documents (LayoutLMv3 Tobacco intra/cross, finance). Feature
whitening (Ledoit--Wolf shrinkage, \citealp{ledoit2004}) is fit on train-ID only. The
base score is kNN-to-bank \citep{sun2022knn}. Settings were included
if the base separability lies in $[0.62, 0.995]$ AUROC (both harm and adaptation must
have headroom). Grids: $\pi\in\{0.01,0.05,0.1,0.5\}$, orders $\{$i.i.d., bursty$\}$,
5 exploration seeds. All headline numbers were re-run once, on the full 96-setting
battery, using five fresh prime seeds never used during development (a strict one-shot
hold-out). Campaign totals: kill $480$ plus its $12$-setting confirmation $480$, grid
$3{,}840$, and ablations $2{,}160$ (together the $6{,}960$ streaming-dynamics cells);
CDC $2{,}304$, bringing all primary campaigns to $9{,}264$ cells; the aggregate-logged
baseline-robustness sweep ($360$ runs, Appendix table); and the one-shot held-out
replication of the full battery ($3{,}840$ cells). Zero cell failures.

\paragraph{Detectors.} static (train-calibrated), frozen oracle (test-ID-calibrated),
AdaODD-style ungated ID bank, OODD-style ungated dictionary, WARDEN (gated; ranks by
base score, decides by two conformal channels at $\alpha/2$, admits by e-BH at
$\delta$), CDC (Section~\ref{sec:cdc}), and the affine-refresh ablation.
Ranking and decisions are decoupled in WARDEN. AUROC is computed on the base ranking
score, so it equals the frozen baseline by construction; the flags come from two
conformal channels at $\alpha/2$ (base and dictionary proximity), each super-uniform
against the frozen reserve, so their union keeps FPR $\le\alpha$. We do not claim the
dictionary channel improves detection. A paired ablation on the full grid shows it does
not: at a matched FPR budget it adds a median $+0.000$ TPR over the base channel alone,
and it leaves WARDEN a median $0.11$ TPR below spending the whole budget on the base
channel (Appendix~\ref{app:c2}). This is the expected consequence of our no-gain
finding, and we report it rather than hide it. WARDEN's contribution is safety, not
detection power: certified admission keeps the dictionary from poisoning the detector,
and the conformal decision holds FPR at a conservative realized mean of $0.057$.

\begin{figure}[t]
\centering
\includegraphics[width=\textwidth]{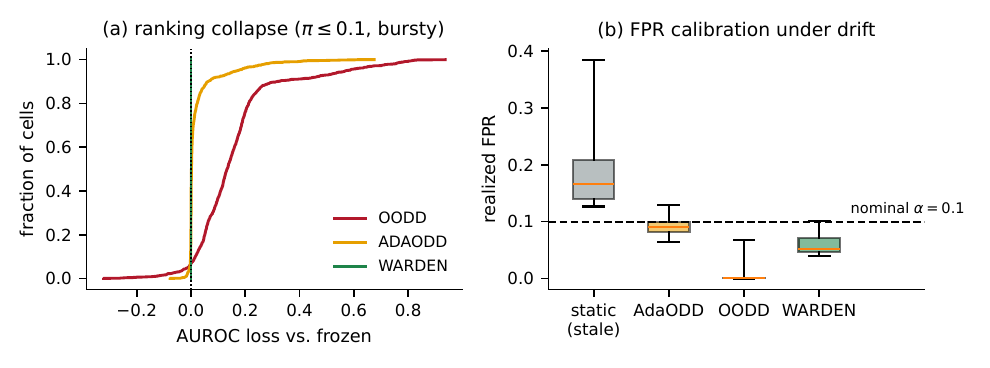}
\caption{Headline outcomes over all bursty cells with $\pi\le0.1$.
(a)~Empirical CDFs of the AUROC loss relative to each detector's own frozen baseline:
OODD-style dictionaries lose ranking quality catastrophically (median $0.13$,
tail beyond $0.9$), AdaODD-style ID banks lose mildly (median $0.002$, consistent with
the theory's small $a(\pi)$ for the mirror mechanism), and WARDEN is a point mass at
zero by construction. (b)~Realized FPR at nominal $\alpha=0.1$ (boxes: IQR, whiskers:
$5$--$95\%$): static stale calibration inflates to $\approx2\times$ nominal; the
ungated detectors' fixed thresholds drift out of control in both directions, OODD's collapses toward zero (it flags almost nothing: a silent failure, since its
AUROC is simultaneously destroyed), AdaODD's scatters widely around nominal, while
WARDEN's realized FPR is the only one consistent with the nominal level throughout
(mean $0.060$; the conformal guarantee is marginal per decision, and per-cell
realizations fluctuate around it, max $0.114$).}
\label{fig:headline}
\end{figure}

\paragraph{Findings (all confirmed on held-out seeds; Figure~\ref{fig:headline}).}
\begin{enumerate}[leftmargin=1.4em,itemsep=1pt,topsep=2pt]
\item Collapse: ungated dictionary $-0.163$ AUROC vs frozen (bursty, $\pi\le
0.1$); impurity $>0.9$; realized FPR uncontrolled in both directions. Held-out
confirmation (fresh primes): \CONFcollapseOODD{}. The collapse is not a tuned-baseline
artifact. Sweeping the admission fraction $q$ over $0.05$, $0.1$, $0.2$, $0.3$, $0.5$
(bursty, $\pi\le0.05$, 12 settings $\times$ 3 seeds), the dictionary is harmed at
every $q$ on 12/12 settings, with mean $\Delta$AUROC rising monotonically
$0.08 \to 0.15 \to 0.29 \to 0.31 \to 0.37$ and the harmed-cell fraction from $61\%$
to $99\%$ (Appendix table). The ID-bank variant's harm is mild at every $q$ (mean
$\Delta$AUROC $\le 0.02$), consistent with the theory, since its admission kernel has
far smaller $a(\pi)$ at low contamination.
\item Threshold law: the admission kernel is strikingly affine. Pooling
$4.3\times10^6$ admission events, the false-admission proportion fits
$q(\rho)=a+b\rho$ with $R^2 = 0.996$--$0.997$ in every encoder family, with slope just
below one ($b = R_0 = 0.944$--$0.947$, $a\approx0.05$), giving predicted
equilibrium $\rho^* = a/(1-b) \approx 1$ (complete poisoning), consistent with measured
impurity $>0.9$. Per-setting fits (not pooled) give median $R^2 = 0.995$ with
$R_0 \in [0.929, 0.959]$ on all $96$ settings, so the fit quality is not a
cross-setting pooling artifact. The $\pi$-conditional coefficients, and the reading of
the tight aggregate slope as a protocol signature rather than an encoder constant, are
in Appendix~\ref{app:exp}. The
predicted $\pi_c$ lands within the pre-registered half-decade tolerance of the empirical
knee on 96/96 knee-settings; impurity trajectories are monotone in $92\%$ of poisoned
cells.
\item Gating: over the $4{,}800$ standard-configuration cells (kill, confirmation,
grid), WARDEN impurity $\le 0.111$ ($\delta = 0.10$) in every cell, and AUROC equal to
frozen everywhere (by construction, hence not a performance claim; the informative
metrics are impurity, realized FPR, and TPR); mean FPR $0.056$ (per-family means
$0.047$--$0.061$; per-cell max $0.121$). The FPR guarantee is marginal per decision
(Proposition~\ref{prop:validity}), and the per-cell maximum is consistent with
binomial fluctuation at the per-cell ID counts.
\item Staleness and CDC: stale FPR $0.194$; CDC certifies $100\%$ of the tested
drift-affected cells at mean FPR $0.042$ with median $67\%$ oracle-TPR retention (near the worst-case
label-free ceiling of Theorem~\ref{thm:impossible}); bursty stress unchanged.
\item Held-out replication: a single one-shot re-run of the full battery
($3{,}840$ cells) on five fresh prime seeds passes every pre-registered hard gate:
collapse reproduces on the gate population (bursty, $\pi\le0.1$ held-out cells:
$\Delta$AUROC $0.114$, $84\%$ of those cells harmed, impurity $0.93$; the collapse
magnitude is seed-sensitive, $0.11$--$0.16$, while the harmed fraction, impurity, and
every guarantee are stable), WARDEN holds (gate population: $\Delta$AUROC $0.000$,
mean FPR $0.059$, impurity $\le 0.091$; over all $3{,}840$ held-out cells, mean FPR
$0.057$, max $0.115$, with final impurity above $0.111$ only in six cells whose
dictionaries hold fewer than ten points), static inflation $1.94\times$, and CDC
certifies $100\%$ of tested cells (FPR $0.043$; retention $0.632$; bursty $100\%$ at
$0.044$).
\item Ablations: conclusions stable across $\delta\in[0.05,0.3]$,
$k\in\{5,10,20\}$, $\alpha\in\{0.05,0.1\}$, calibrator exponent, and screening
threshold.
\end{enumerate}

\section{Limitations and scope}\label{sec:limitations}

(i) Assumption~\ref{ass:kernel} posits a one-dimensional state. The affine fit is
excellent empirically but the reduction is a modelling step, and richer state
(e.g.\ cluster structure of the poison) is future work. (ii)
Theorem~\ref{thm:cdc} assumes an exchangeable calibration window.
Proposition~\ref{prop:block} extends the certificate to $L$-block-dependent windows at
effective size $n/L$, but that worst-case slack is loose (the i.i.d.-calibrated CDC
already certifies $100\%$ of bursty cells). A tight dependent-window certificate
(block-bootstrap quantiles matched to the burst length) is the principal open
theoretical problem. (iii) CDC's power loss is real (median $\sim\!33$--$37\%$ of oracle
TPR at $\pi\le0.1$). Theorem~\ref{thm:impossible} bounds how much of it is removable in
the worst case (retention above $\kappa$ is impossible while keeping the certificate),
but on benign instances part of the loss is finite-sample slack, and sharper estimates
of the confusable mass (e.g.\ via mixture-proportion irreducibility) could recover some
of it. We do not claim CDC is instance-optimal. Our guarantees are also per-batch rather than time-uniform. An anytime-valid
version of the admission guarantee via stopped e-BH \citep{stoppedebh} is open (their theory names FDR control under global dependence as unresolved, which is exactly
the deployment filtration here). (iv) Adaptation buys no AUROC on well-whitened
features. If
ranking quality is the goal, adaptation is the wrong tool on these representations.
(v) All experiments use cached features. End-to-end retraining dynamics are out of
scope. (vi) Corollary~\ref{cor:adversarial} is stress-tested only with the feature-space
tail-mimicry attack of Appendix~\ref{app:attack}, which instantiates its threat model
but does not optimize the attack; gradient-based input-space poisoning in the style of
\citet{ttapoison} against the admission certificate is untested. (vii) The empirical evidence spans
vision, text, and document encoders. Deployments in other domains with the same
structural ingredients (transaction fraud, intrusion detection, clinical monitoring)
are in scope of the theory but untested here, and the kernel coefficients $a(\pi)$,
$b(\pi)$ are domain-specific quantities that must be re-measured before any threshold
prediction is trusted.

\section{Reproducibility}
All campaigns are driven by a resumable, single-command pipeline (settings discovery,
grids, ablations, held-out confirmation, analysis, figure and table generation).
Per-cell JSON artifacts and verdict files are emitted for every number in this paper.
The code and artifacts will be released publicly upon acceptance.

\acks{This work was carried out independently and received no specific grant from any
funding agency in the public, commercial, or not-for-profit sectors.}

\newpage
\appendix

\section{Proofs for Section~\ref{sec:theory}}\label{app:theory}

\subsection{Derivation of the recursion \texorpdfstring{\eqref{eq:sa}}{(SA)}}
With $m_t=m_{t-1}+w_t$, $n_t=n_{t-1}+a_t$, and $\rho_t=m_t/n_t$,
\[
\rho_t-\rho_{t-1}
=\frac{m_{t-1}+w_t}{n_{t-1}+a_t}-\frac{m_{t-1}}{n_{t-1}}
=\frac{n_{t-1}w_t-m_{t-1}a_t}{n_{t-1}(n_{t-1}+a_t)}
=\frac{a_t}{n_t}\Bigl(\frac{w_t}{a_t}-\rho_{t-1}\Bigr),
\]
where the last step uses $m_{t-1}=\rho_{t-1}n_{t-1}$ and $n_t=n_{t-1}+a_t$. Setting
$\gamma_t=a_t/n_t$ and $\xi_t=w_t/a_t-q(\rho_{t-1})$ yields \eqref{eq:sa} exactly, with
$h(\rho)=q(\rho)-\rho$. Under Assumption~\ref{ass:kernel} the admission size $a_t$
(hence $n_t$ and $\gamma_t$) is $\mathcal F_{t-1}$-measurable, so
$\mathbb E[\gamma_t\xi_t\mid\mathcal F_{t-1}] = \gamma_t\,
\mathbb E[\xi_t\mid\mathcal F_{t-1}] = 0$ on $\{a_t>0\}$: the perturbation is a bounded
martingale difference with predictable envelope $\gamma_t \le a_{\max}/n_{t-1}$. (Were
$a_t$ not predictable, the drift would acquire the bias
$\mathrm{Cov}(\gamma_t,\, w_t/a_t \mid \mathcal F_{t-1})$ and the mean field would be
the admission-weighted kernel $\bar q$; see the remark after
Assumption~\ref{ass:kernel}.) \hfill\BlackBox

\subsection{Proof of Theorem~\ref{thm:converge}}
Write the recursion \eqref{eq:sa}. On $\{n_t\to\infty\}$: since $a_t\le a_{\max}$ and
$n_t \ge n_0 + \#\{\text{admitting batches} \le t\}$, we have
$\gamma_t \le a_{\max}/n_{t-1}$ with $\sum_t \gamma_t = \infty$ (each admitting batch
contributes at least $1/n_t$ and $n_t$ grows at most linearly) and
$\sum_t \gamma_t^2 < \infty$ ($\gamma_t = O(1/t)$ along admitting batches). By the
derivation above, $\gamma_t\xi_t$ is a bounded martingale-difference perturbation with
predictable step. $h$ is Lipschitz on
$[0,1]$ with $h(0)=q(0)\ge0$ and $h(1)=q(1)-1\le 0$, so the flow of
$\dot\rho = h(\rho)$ leaves $[0,1]$ invariant and its chain-recurrent set is the zero
set of $h$ (scalar flow). By the ODE method for stochastic approximation
\citep[Prop.~4.1, Cor.~6.6]{benaim1999}, whose deterministic-gain statements apply
pathwise here because the steps $\gamma_t$ are predictable with
$\sum\gamma_t=\infty$, $\sum\gamma_t^2<\infty$ a.s.\ on the admission-accumulation
event, so the interpolated process is an asymptotic pseudotrajectory of the flow, $\rho_t$ converges a.s.\ to a connected
internally chain-transitive set, i.e.\ to the zero set; isolated zeros give convergence
to a single zero. Non-convergence to a linearly unstable zero $\rho^u$ (where
$h'(\rho^u) = q'(\rho^u)-1 > 0$) follows from \citet{pemantle2007} (Theorem 2.9,
originally Pemantle 1990) provided the conditional noise variance is bounded
below in a neighborhood of $\rho^u$, exactly the scope of
Assumption~\ref{ass:rate}; boundary zeros where the noise floor is inactive (e.g.\
$\rho^u=0$ reached from a clean bank with $q(0)=0$) are excluded from the claim, as the
theorem statement says. \hfill\BlackBox

\subsection{Proof of Corollary~\ref{cor:affine}}
For $q = \min\{a+b\rho, 1\}$: if $b<1$, $h(\rho) = a-(1-b)\rho$ on the unsaturated
range, strictly decreasing with unique zero $\rho^* = a/(1-b)$ (or, if
$a/(1-b) > 1$, $h>0$ up to the saturation region where $h(\rho)=1-\rho$, giving the
zero at $1$); $h' <0$ so the zero is stable and Theorem~\ref{thm:converge} applies. If
$b\ge1$: on the unsaturated range $h(\rho) = a + (b-1)\rho \ge a > 0$ (at $b=1$ exactly,
$h \equiv a > 0$ there); on the saturated range
$h = 1-\rho \ge 0$ with equality only at $\rho=1$; thus $h>0$ on $[0,1)$, the flow is
strictly increasing, the only chain-recurrent point is $1$, and $\rho_t\to1$ a.s.
\hfill\BlackBox

\subsection{Proof of Proposition~\ref{prop:fold}}
At most three crossings. $h'(\rho)=\partial_\rho q-1$; since $\partial_\rho q$ is
strictly unimodal (hypothesis (ii)), $h'$ has at most two sign changes
($-\!\to\!+\!\to\!-$), so $h$ has at most three zeros, and by (i) ($h(0)>0$, $h(1)<0$)
the number of zeros is odd: one or three.
Bistable window is a nonempty interval. By (iii), $h(\rho;\theta)$ is strictly
increasing in $\theta$ pointwise. Define
$\theta_- := \inf\{\theta:\ h(\cdot;\theta)\text{ has three zeros}\}$ and
$\theta_+ := \sup\{\cdot\}$; the three-zero set is an interval because increasing
$\theta$ raises $h$, which can destroy the lower zero pair only once (the middle
negative band of $h$ shrinks monotonically in $\theta$), pair creation at
$\theta_-$ and destruction at $\theta_+$ are each one-shot. Hypothesis (iv) makes the
set nonempty ($\theta^*$ has a steep crossing region between the endpoint regimes,
giving three zeros for some $\theta$) and places the unique low (high) crossing at the
sweep ends, and (v) with the implicit function theorem makes each boundary a
non-degenerate saddle-node with $\theta_-<\theta_+$ strictly (a coincident tangency
would force $\partial^2_{\rho\rho}q=0$ at the inflection, excluded by (v)).
Hysteresis. For $\theta$ slightly below $\theta_+$ the flow started on the low
branch stays on it (it is separated from the high branch by the middle unstable zero);
at $\theta_+$ the low branch is annihilated in the fold and the state jumps to the high
branch; sweeping down, the high branch survives until $\theta_-$: the attained
equilibrium is direction-dependent on $(\theta_-,\theta_+)$. \hfill\BlackBox

\subsection*{Proof of Proposition~\ref{prop:fifo} (uniform eviction)}
With $n_t \equiv N$: admit $a_t$ points (wrong count $w_t$), evict a uniformly random
$a_t$-subset of the $N$ incumbents, whose conditional mean impurity is exactly
$\rho_{t-1}$. Hence
$\mathbb E[\rho_t\mid\mathcal F_{t-1}]
= \rho_{t-1} + \frac{a_t}{N}\bigl(q(\rho_{t-1}) - \rho_{t-1}\bigr)$ exactly, and for the
affine kernel $\mathbb E[\rho_t] - \rho^* = (1 - \gamma(1-b))(\mathbb E[\rho_{t-1}] -
\rho^*)$: geometric contraction with no bias term. For the variance, the conditional
mean is affine in $\rho_{t-1}$ alone (this is where uniform eviction is used a second
time), so by the conditional-variance decomposition
$v_t \le (1-\gamma(1-b))^2 v_{t-1} + \gamma^2\sigma^2_{\max}$ with
$\sigma^2_{\max} = \sup\mathrm{Var}(w_t/a_t - \rho^{\mathrm{ev}}_t\mid\mathcal F_{t-1})
\le 1$, which converges to
$\gamma^2\sigma^2_{\max}/(1-(1-\gamma(1-b))^2) \le
\gamma\,\sigma^2_{\max}/(2(1-b)-\gamma(1-b)^2)$. Under literal FIFO,
$\rho^{\mathrm{ev}}$ is the lagged block impurity; writing the same recursion yields a
delay equation with identical unique fixed point and an additional
$O(\gamma)$-amplitude term, which we do not pursue. \hfill\BlackBox

\subsection{Proof of Lemma~\ref{lem:ebh} and Theorem~\ref{thm:gating}}
Threshold structure of e-BH. With $K$ hypotheses at level $\delta$, e-BH rejects
the $k^*$ largest e-values where $k^* = \max\{k : e_{(k)} \ge K/(\delta k)\}$. Every
admitted point therefore has $e_i \ge K/(\delta a_t) \ge 1/\delta$ (as $a_t \le K$).
Since $e = a p^{a-1}$ is decreasing in $p$, admission requires
$p_i \le \kappa(\delta) = (a\delta)^{1/(1-a)}$.
Per-point wrong-admission probability. Condition on the realized reserve $R$
(scores $r_{(1)}\le\dots\le r_{(m)}$, frozen scorer). A true-ID point is admitted only
if $p_i \le \kappa$, i.e.\ its score exceeds the top $\lceil\kappa(m+1)\rceil-1$ reserve
scores; conditionally on $R$ this event has probability $\nu(R) := \Pid\bigl(s >
r_{(m-\lceil\kappa(m+1)\rceil+1)}\bigr)$, a fixed number with
$\mathbb E[\nu(R)] \le \kappa$ by exchangeability. For the high-probability version,
note that for any $u$, $\{\nu(R) > u\}$ is exactly the event that fewer than
$\lceil\kappa(m+1)\rceil$ of the $m$ (i.i.d.) reserve points fall in the top-$u$ mass
of $\Pid$, i.e.\ $\{\mathrm{Bin}(m,u) < \lceil\kappa(m+1)\rceil\}$, whose probability
is decreasing in $u$; by the definition of $\bar\kappa$ as the largest $u$ at which
this probability still reaches $\eta$,
$\Pr\bigl(\nu(R) > \bar\kappa\bigr) \le \eta$ exactly, no regime conditions.
(The familiar Chernoff closed form $\bar\kappa \le \kappa +
\sqrt{c\,\kappa\log(1/\eta)/m}$ holds with $c=4$ once
$m\kappa \gtrsim \log(1/\eta)$; at our operating point $m\kappa \approx 9$ we use the
exact value.)
Intensity bound. On the event $\{\nu(R)\le\bar\kappa\}$ (probability
$\ge1-\eta$, one draw, all $t$), linearity of conditional expectation over the at most
$K$ true-ID points of batch $t$ gives $\mathbb E[w_t\mid\mathcal F_{t-1}] \le
K\bar\kappa = \bar W$, regardless of the dictionary state, the admission
p-values never involve the dictionary, and regardless of within-batch dependence
(the bound is a union of marginals). This proves the lemma.
Theorem. (i) $M_t := m_t - \sum_{s\le t}\mathbb E[w_s\mid\mathcal F_{s-1}]$ is a
martingale with increments bounded by $K$, so $M_t/t\to0$ a.s.\ (SLLN for bounded
martingale differences), giving $\limsup m_t/t \le \bar W$. (ii) On
$\{\liminf n_t/t \ge r\}$: $\limsup \rho_t = \limsup m_t/n_t \le \bar W/r$. (iii) The
gated admission-weighted kernel obeys $\bar q_{\mathrm{gated}}(\rho) =
\mathbb E[w_t\mid\cdot]/\mathbb E[a_t\mid\cdot] \le \bar W/(\bar W + r_{\mathrm{ood}})$
where $r_{\mathrm{ood}}$ is the true-OOD admission intensity; the right side does not
depend on $\rho$, so $h$ gains no reinforcement term and has no supercritical branch.
\hfill\BlackBox

\begin{remark}\label{rem:fdr-not-enough}
Per-batch FDR control alone cannot yield Theorem~\ref{thm:gating}: e-values equal to
$K/\delta$ with probability $\delta/K$ (else $0$) on pure-ID batches satisfy
$\mathbb E[\mathrm{FDP}]\le\delta$ yet admit only wrong points, driving
$\rho_t\to1$. The theorem's force comes from the severed feedback (admission evidence
frozen) plus the absolute threshold $e\ge1/\delta$; e-BH additionally keeps the
per-batch false-admission proportion at $\delta$ in expectation
(Proposition~\ref{prop:validity} controls the decision; this controls the
admission mix). Measured impurity: $\le0.111$ over all $6{,}960$ cells, far
below the a-priori ceiling $\bar W/r$.
\end{remark}

\subsection{Proof of Corollary~\ref{cor:adversarial}}
Fix a batch. The scorer and the reserve are $\mathcal F_{t-1}$-measurable, so for each
true-ID point in the batch the joint law of its score and the reserve scores is
exchangeable no matter what else the batch contains. Its conformal p-value is therefore
super-uniform conditionally on $\mathcal F_{t-1}$, and the contaminated points enter
neither this joint law nor the frozen evidence channel, so
Proposition~\ref{prop:validity} is unchanged. For Lemma~\ref{lem:ebh}, every e-BH
admission still requires $p_i\le\kappa(\delta)$, a deterministic threshold argument
the adversary cannot touch. The number of wrong admissions in the batch is at most the
number of true-ID points with $p_i \le \kappa$, and the conditional expectation of
that count is bounded by $K\bar\kappa$ on the reserve event exactly as in the lemma's
proof. Which of these points e-BH selects may depend on the adversary, but the bound
counts all of them, so it is adversary-free. Theorem~\ref{thm:gating} uses only the
per-batch bound and the admission-accrual rate $r$, so its proof goes through verbatim.
\hfill\BlackBox

\section{Proofs for Section~\ref{sec:cdc}}\label{app:cdc}

\subsection{Proof of Theorem~\ref{thm:cdc}}
By the one-sided DKW inequality with Massart's constant \citep{massart1990}, with
probability $\ge 1-\eta/2$,
$\sup_t \bigl(\hat F_{\mathrm{mix}}(t) - \Fmix(t)\bigr) \le \eps_n - 1/n$ (validity
needs only this direction; the two-sided event used for Theorem~\ref{thm:power} costs
$\eta$). On that event, by construction of $\hat\tau$,
$\Fmix(\hat\tau) \ge \hat F_{\mathrm{mix}}(\hat\tau) - \eps_n \ge
1 - \alpha(1-\hat\pi_{\mathrm{up}})$, i.e.\
$\Pmix(s>\hat\tau) \le \alpha(1-\hat\pi_{\mathrm{up}})$. Since
$\Pmix = (1-\pi)\Pid + \pi\Pood \ge (1-\pi)\Pid$ pointwise,
$\Pid(s>\hat\tau) \le \Pmix(s>\hat\tau)/(1-\pi)$. On the event
$\{\hat\pi_{\mathrm{up}} \ge \pi\}$ (probability $\ge 1-\eta_1$),
$\alpha(1-\hat\pi_{\mathrm{up}})/(1-\pi) \le \alpha$. Union bound. Note the argument
uses only nonnegativity of the OOD component, no dominance, and that $\hat\tau$
is computed on past data, so the evaluated point is independent of it under the i.i.d.\
window assumption. \hfill\BlackBox

\subsection{Proof of Theorem~\ref{thm:storey}}
(a) Work conditionally on $R_{\mathrm{tr}}$ throughout; the window is i.i.d.\
given $R$. Then $\mathbb E[\#\{p_i>\lambda\}\mid R]/n = (1-\pi)\Pid(p>\lambda\mid R) +
\pi\kappa_\lambda(R) = (1-\pi)[(1-\lambda) - \eps_{\mathrm{dr}}(R)] +
\pi\,\kappa_\lambda(R)$, hence
\[
\mathbb E[\hat\pi_\lambda \mid R]
= \pi + \frac{(1-\pi)\eps_{\mathrm{dr}}(R) - \pi\kappa_\lambda(R)}{1-\lambda}
\;\ge\; \pi \quad\text{under (A2)}.
\]
Given $R$, the indicators $\ind{p_i>\lambda}$ are i.i.d.\ (they share only the
fixed reserve), so Hoeffding applies conditionally:
$\hat\pi_\lambda \ge \mathbb E[\hat\pi_\lambda\mid R] -
\frac{1}{1-\lambda}\sqrt{\log(2/\eta)/(2n)}$ with probability $\ge 1-\eta/2$; the Step-1
slack gives $\Pr(\hat\pi_{\mathrm{up}}<\pi \mid R)\le\eta/2$. $\pi_\dagger(R)$ is (A2)
solved for $\pi$.
(b) With $\lambda = k_0/(m+1)$ on the grid and continuous scores, an
exchangeable point's conformal p-value satisfies $\Pr(p \le \lambda) = \lambda$
exactly (marginally over $(R, s)$), so dominance of test-ID over train-ID scores, which makes the test-ID p-value stochastically smaller than the exchangeable one
against the same reserve, pointwise in $R$, gives
$\mathbb E_R[\Pid(p>\lambda\mid R)] \le 1-\lambda$, i.e.\
$\bar\eps_{\mathrm{dr}} := \mathbb E_R[\eps_{\mathrm{dr}}(R)] \ge 0$. Both
$\Pid(p>\lambda\mid R)$ and $\kappa_\lambda(R)$ are tail probabilities at the reserve's
$\lceil(1-\lambda)(m+1)\rceil$-th order statistic, so DKW on the reserve empirical CDF
bounds their fluctuations by $\eps_m$ each with probability $\ge 1-2\eta_R$; the
population margin condition then implies (A2), and the total failure probability is
$\eta/2 + 2\eta_R$ by a union bound. \hfill\BlackBox

\subsection{Proof of Theorem~\ref{thm:power}}
On the DKW event, by construction $\hat F_{\mathrm{mix}}(\hat\tau) \ge
1-\alpha(1-\hat\pi_{\mathrm{up}})+\eps_n - 1/n$, so the level of $\hat\tau$ in $\Fmix$
lies in $\bigl[1-\alpha(1-\hat\pi_{\mathrm{up}}),\;
1-\alpha(1-\hat\pi_{\mathrm{up}})+2\eps_n\bigr]$ (consistent with the validity proof;
the $1/n$ is inside $\eps_n$). The level of $\tau^*$ in $\Fmix$ is
$\Fmix(\tau^*) = (1-\pi)(1-\alpha) + \pi\Food(\tau^*) \ge (1-\pi)(1-\alpha)$, using
only $\Food(\tau^*)\ge 0$, the bound must be from below to upper-bound the
gap. Hence the level gap satisfies
\[
\Fmix(\hat\tau)-\Fmix(\tau^*) \;\le\;
\bigl[1-\alpha(1-\hat\pi_{\mathrm{up}})+2\eps_n\bigr] - (1-\pi)(1-\alpha)
= \alpha\hat\pi_{\mathrm{up}} + (1-\alpha)\pi + 2\eps_n .
\]
The density lower bound $\kappa$ on $[\tau^*,\hat\tau]$ converts the level gap to
$\hat\tau - \tau^* \le (\alpha\hat\pi_{\mathrm{up}} + (1-\alpha)\pi + 2\eps_n)/\kappa$,
and the OOD density upper bound converts the threshold gap to the TPR gap. The drift
floor in the statement follows by substituting the systematic part
$\bar\eps_{\mathrm{dr}}/(1-\lambda)$ of $\hat\pi_{\mathrm{up}}$ from
Theorem~\ref{thm:storey}(b). \hfill\BlackBox

\subsection{Proof of Proposition~\ref{prop:block}}
Write the window as independent blocks $Z^{(1)},\dots,Z^{(B)}$ of lengths $l_j\le L$,
$\sum_j l_j = n$, each with marginal point law $\Fmix$. Independence across blocks is
essential: exchangeability alone admits de Finetti mixtures for which no such
concentration holds.
(i) Pointwise deviation. For fixed $t$, $\hat F_{\mathrm{mix}}(t) = \sum_j
(l_j/n)\,\hat F^{(j)}(t)$ is a weighted average of $B$ independent $[0,1]$-valued
variables with mean $\Fmix(t)$ and weights $l_j/n$; Hoeffding gives
$\Pr[|\hat F_{\mathrm{mix}}(t)-\Fmix(t)|>\epsilon]\le
2\exp\bigl(-2\epsilon^2 n^2/\textstyle\sum_j l_j^2\bigr) =
2e^{-2 n_{\mathrm{eff}}\epsilon^2}$.
(ii) Uniform deviation via a quantile grid. Choose $k$ grid points $t_1 < \dots
< t_k$ with $\Fmix(t_i) = i/(k{+}1)$. A union bound gives all $k$ pointwise deviations
$\le\epsilon$ with probability $\ge 1-2k\,e^{-2n_{\mathrm{eff}}\epsilon^2}$;
monotonicity of $\hat F_{\mathrm{mix}}$ and $\Fmix$ then extends the bound between grid
points at cost $1/(k{+}1)$:
$\sup_t|\hat F_{\mathrm{mix}}-\Fmix| \le \epsilon + 1/k$ with probability $\ge
1-2(k{+}1)e^{-2n_{\mathrm{eff}}\epsilon^2}$, which is the stated
$\eps(n_{\mathrm{eff}})$ at $\epsilon=\sqrt{\log(2(k{+}1)/\eta)/(2n_{\mathrm{eff}})}$.
(A naked McDiarmid argument would mis-center: it controls deviation from
$\mathbb E[\sup]$, which is itself $\Theta(n_{\mathrm{eff}}^{-1/2})$.)
(iii) Storey statistic. $\hat\pi_\lambda$ is a length-weighted average of block
means of $\ind{p_i>\lambda}$ (conditionally on the reserve, blocks remain independent);
the same weighted Hoeffding applies at $n_{\mathrm{eff}}$. Substituting
$\eps(n_{\mathrm{eff}})$ for the window terms of
Theorems~\ref{thm:cdc}--\ref{thm:storey}, the quantile and
conservativity arguments are unchanged. \hfill\BlackBox

\section{Proof of Theorem~\ref{thm:impossible}}\label{app:impossible}

Construction and dominance. Let $P$ be continuous with $\tau_B=P^{-1}(1-\alpha)$,
so $\bar P(\tau_B):=P(s>\tau_B)=\alpha$. Let $Q=\mathcal L(s\mid s>\tau_B)$, i.e.
$\bar Q(t)=\bar P(t)/\alpha$ for $t\ge\tau_B$ and $\bar Q(t)=1$ for $t<\tau_B$. Set
$P'=(1-\pi)P+\pi Q$, with tail $\bar P'=(1-\pi)\bar P+\pi\bar Q$. For $t\ge\tau_B$,
$\bar P'(t)=\bar P(t)\bigl[(1-\pi)+\pi/\alpha\bigr]\ge\bar P(t)$ (as $\alpha\le1$); for
$t<\tau_B$, $\bar P'(t)=(1-\pi)\bar P(t)+\pi\ge\bar P(t)$. Hence $P'\succeq P$: $P'$ is a
legitimate upward drift of the ID law.

Indistinguishability. World $A$ (ID $=P'$, no contamination) has stream tail
$\bar P'$. World $B$ (ID $=P$, contamination fraction $\pi$ from $Q$) has stream tail
$(1-\pi)\bar P+\pi\bar Q=\bar P'$. The stale reserve is drawn from $P$ in both. Thus the
observable law (window $+$ reserve) is identical, and any label-free $\mathcal A$
produces $\hat\tau$ with a common law $\mu$ across the two worlds.

The inclusion. Fix any threshold value $\tau$.
Case $\tau<\tau_B$: there $\bar Q(\tau)=1$ and $\bar P(\tau)\ge\alpha$, so
$\mathrm{FPR}_A(\tau)=\bar P'(\tau)=(1-\pi)\bar P(\tau)+\pi \ge (1-\pi)\alpha+\pi
>\alpha$ for every $\pi\in(0,1)$; thus $\{\mathrm{FPR}_A\le\alpha\}$ forces
$\hat\tau\ge\tau_B$.
Case $\tau\ge\tau_B$: $\bar P'(\tau)=\bar Q(\tau)\,[\pi+\alpha(1-\pi)]$ (using
$\bar P(\tau)=\alpha\bar Q(\tau)$), so
$\mathrm{FPR}_A(\tau)\le\alpha \iff \bar Q(\tau)\le\alpha/[\pi+\alpha(1-\pi)]=\kappa$.
Since the world-$B$ outliers are exactly $Q$, $\mathrm{TPR}_B(\tau)=\bar Q(\tau)$. Hence
$\{\mathrm{FPR}_A(\hat\tau)\le\alpha\}\subseteq\{\hat\tau\ge\tau_B\}\cap
\{\bar Q(\hat\tau)\le\kappa\}=\{\mathrm{TPR}_B(\hat\tau)\le\kappa\}$, almost surely under
$\mu$.

Conclusion. Taking $\mu$-probabilities,
$\Pr_A[\mathrm{FPR}\le\alpha]\le\Pr_B[\mathrm{TPR}\le\kappa]$; the contrapositive is the
stated $(1-\beta)$ bound. The labelled oracle sets $\tau=P'^{-1}(1-\alpha)$ in world $A$
(FPR $=\alpha$) and $\tau=\tau_B$ in world $B$ (FPR $=\alpha$, TPR $=\bar Q(\tau_B)=1$),
achieving both objectives because labels break the tie the unlabelled law cannot.
\hfill\BlackBox

\begin{remark}
The ceiling is tight for this pair in the population limit: the procedure that sets
$\hat\tau=P'^{-1}(1-\alpha)$, computable from the stream law, hence approachable by
any consistent quantile estimator as $n\to\infty$, is valid in $A$ and attains
$\mathrm{TPR}_B=\bar Q(\hat\tau)=\kappa$ exactly. The construction also pins down the
role of condition (A2) in Theorem~\ref{thm:storey}: here the OOD leakage above any
$\lambda$ is maximal (outliers sit in the ID tail), so (A2) fails, precisely the
regime where CDC must, and does, forfeit power rather than validity.
\end{remark}

\section{Additional experimental detail}\label{app:exp}

\paragraph{Measured admission kernel (validates Assumption~\ref{ass:kernel}).}
Pooling all OOD-dictionary admission events by encoder family and fitting the
false-admission proportion $q(\rho)=a+b\rho$ by weighted least squares over $12$ impurity
bins:
\begin{center}\small
\begin{tabular}{lcccc}
\toprule
family & $a$ & $b=R_0$ & $R^2$ & \#admissions \\
\midrule
vision (trained) & $0.045$ & $0.947$ & $0.996$ & $2.1\times10^6$ \\
vision-foundation & $0.046$ & $0.947$ & $0.997$ & $8.0\times10^5$ \\
document & $0.066$ & $0.944$ & $0.997$ & $7.1\times10^5$ \\
text & $0.045$ & $0.947$ & $0.996$ & $6.7\times10^5$ \\
\midrule
all & $0.048$ & $0.947$ & $0.997$ & $4.3\times10^6$ \\
\bottomrule
\end{tabular}
\end{center}
The near-perfect affine fit justifies the one-dimensional mean field of
Section~\ref{sec:theory}; the equilibrium $a/(1-b)\approx1$ predicts complete
poisoning, matching measured impurity $>0.9$.

Fitting each of the 96 settings individually gives per-setting $R^2$ with median
$0.995$, tenth percentile $0.992$, minimum $0.916$, and $R^2 \ge 0.95$ on $95$ of
$96$ settings, with the per-setting slope in $[0.929, 0.959]$ on all $96$. The fit
quality is therefore not an artifact of cross-setting pooling.

\paragraph{Where the tight slope comes from.} The pooled and per-setting fits average
over the $\pi$ grid and the burst phases, so they estimate the admission-weighted
kernel $\bar q$ that drives recursion~\eqref{eq:sa}. Conditioning on $\pi$ decomposes
the same law into the pair $(a(\pi), b(\pi))$ the theory consumes. For the OOD
dictionary (same fit protocol), $(a,b) = (0.94, 0.03)$ at $\pi=0.01$, $(0.51, 0.42)$
at $0.05$, $(0.17, 0.77)$ at $0.1$, and $(0.02, 0.90)$ at $0.5$. At low contamination
the collapse is intercept-driven, pure-ID bursts force wrong admissions regardless of
the bank state. At high contamination it is slope-driven, reinforcement through the
contrast score. The concentration of the aggregate slope near one is thus a signature
of the fixed-fraction, bank-relative admission protocol, which pins the composition
of admitted points close to the bank's current composition, and not an encoder-level
constant. Two controls support this reading. The mirror ID-bank rule, the same
protocol pointed at the opposite tail, has aggregate slope $1.006$ with
$\pi$-conditional slopes $0.04$, $0.56$, $0.96$ at $\pi = 0.05, 0.1, 0.5$ (at
$\pi=0.01$ its impurity never leaves a neighborhood of zero, so there is no impurity
range to fit). And the severed-feedback gate of Section~\ref{sec:gating}, whose
admission evidence never reads the bank, confines admission-time impurity to
$[0, 0.111]$ over $4.0\times10^5$ admissions with mean false-admission proportion
$0.002$ and correlation $0.01$ between wrongness and impurity. With the evidence
channel frozen there is no reinforcement coordinate left to measure. The theorems
consume only the measured $(a(\pi), b(\pi))$, no universality is assumed anywhere,
and the operational $\pi_c$ is computed from these conditional coefficients.

\paragraph{Baseline-robustness sweep (rebuts baseline tuning).} Sweeping the admission
fraction of both ungated baselines over $q\in\{0.05,0.1,0.2,0.3,0.5\}$ (12 settings,
bursty, $\pi\in\{0.01,0.05\}$, 3 seeds):
\begin{center}\small
\begin{tabular}{lcc|cc}
\toprule
& \multicolumn{2}{c|}{OOD dictionary (OODD-style)} & \multicolumn{2}{c}{ID bank (AdaODD-style)} \\
$q$ & mean $\Delta$AUROC & frac.\ harmed & mean $\Delta$AUROC & frac.\ harmed \\
\midrule
$0.05$ & $0.082$ & $0.611$ & $-0.001$ & $0.014$ \\
$0.10$ & $0.151$ & $0.792$ & $0.002$ & $0.056$ \\
$0.20$ & $0.286$ & $0.917$ & $0.010$ & $0.153$ \\
$0.30$ & $0.307$ & $0.917$ & $0.014$ & $0.139$ \\
$0.50$ & $0.367$ & $0.986$ & $0.019$ & $0.139$ \\
\bottomrule
\end{tabular}
\end{center}
Dictionary collapse persists at every admission fraction on 12/12 settings and worsens
monotonically with $q$, as the theory predicts (larger $q$ = larger effective step
$\gamma$ and larger $a(\pi)$); the harm is a property of ungated admission, not of a
particular hyper-parameter choice.

\subsection{Does the dictionary decision channel add power? (No.)}\label{app:c2}
WARDEN decides with two conformal channels at $\alpha/2$, a base channel and a
dictionary-proximity channel (C2). We test whether C2 adds detection power with a
paired ablation on the full grid ($3{,}840$ cells): on each stream WARDEN is run
against two variants that share its admission exactly (hence hold the same dictionary
at every step) and differ only in the decision rule, base-at-$\alpha$ (the whole budget
on the base channel, C2 off) and base-at-$\alpha/2$ (the base channel alone).

\begin{center}\small
\setlength{\tabcolsep}{5pt}
\resizebox{\textwidth}{!}{%
\begin{tabular}{lccccc}
\toprule
$\pi$ & C2 fires & C2 above base ROC$^\dagger$ & C2 median $\Delta$TPR & WARDEN $-$ base-at-$\alpha$ (med) & FPR WARDEN/base-at-$\alpha$ \\
\midrule
$0.01$ & $9\%$  & $8\%$  & $+0.000$ & $-0.109$ & $0.05/0.10$ \\
$0.05$ & $14\%$ & $20\%$ & $+0.002$ & $-0.114$ & $0.06/0.10$ \\
$0.10$ & $17\%$ & $23\%$ & $+0.003$ & $-0.112$ & $0.06/0.10$ \\
$0.50$ & $31\%$ & $28\%$ & $+0.008$ & $-0.109$ & $0.06/0.10$ \\
\bottomrule
\end{tabular}}
\end{center}
$^\dagger$ fraction of the cells where C2 fired in which its TPR-per-FPR exceeds the
base ROC slope, i.e.\ C2 detected outliers the base score ranks below the threshold.

Three facts follow. (i) C2 is inert on well-whitened features: WARDEN and
base-at-$\alpha/2$ have the same mean TPR to three decimals ($0.626$ vs $0.623$); C2
fires in $18\%$ of cells overall and, when it fires, sits above the base ROC in only
$23\%$ of them, for a paired median gain of $+0.000$ TPR. (ii) At a matched FPR budget
WARDEN detects a median $0.11$ TPR less than base-at-$\alpha$, because halving the base
budget costs more than the dictionary channel returns. (iii) The conservatism is a
feature, not a bug: WARDEN's realized FPR (mean $0.057$, max $0.121$) stays well below
base-at-$\alpha$ (mean $0.099$, above the nominal level in $44\%$ of cells, max
$0.158$). We therefore make no detection-gain claim for the dictionary channel. This is
the expected consequence of the no-gain finding and reinforces the paper's thesis: an
adaptive OOD dictionary does not improve detection on well-whitened features even when
it is safely gated. A deployment that prioritizes power over conservative FPR can use
the base channel at level $\alpha$ ($+0.13$ median TPR at the cost of per-cell FPR
fluctuation above $\alpha$); WARDEN as reported takes the conservative operating point.

\subsection{Adversarial stress test of the certificates}\label{app:attack}
Corollary~\ref{cor:adversarial} asserts that no placement of the contaminated points
can break the admission certificate or the decision-level FPR. We instantiate its
threat model with a gray-box tail-mimicry attack. The adversary sees the features, the
reserve, and the detector's code, controls only the contaminated points (never the ID
points or the reserve, matching the corollary's premise), and moves every outlier
toward its nearest ID-reserve neighbour in whitened feature space,
$x' = (1-\eps)\,x + \eps\, r_{\mathrm{nn}}(x)$, with strength
$\eps \in \{0, 0.25, 0.5, 0.75, 0.9\}$. As $\eps \to 1$ the contamination approaches
the confusable ID-tail law of Theorem~\ref{thm:impossible}, the least-favorable
direction. Twelve settings, $\pi \in \{0.05, 0.1\}$, bursty order, three seeds, 72
cells per strength (360 total, zero failures).

\begin{center}\small
\setlength{\tabcolsep}{5pt}
\begin{tabular}{lcccc}
\toprule
$\eps$ & WARDEN FPR mean/max & WARDEN impurity max & WARDEN TPR & ungated impurity \\
\midrule
$0$    & $0.058$ / $0.112$ & $0.008$ & $0.615$ & $0.913$ \\
$0.25$ & $0.049$ / $0.096$ & $0.000$ & $0.216$ & $0.917$ \\
$0.5$  & $0.045$ / $0.061$ & $0.000$ & $0.041$ & $0.920$ \\
$0.75$ & $0.045$ / $0.061$ & $0.000$ & $0.004$ & $0.923$ \\
$0.9$  & $0.045$ / $0.061$ & $0.000$ & $0.004$ & $0.928$ \\
\bottomrule
\end{tabular}
\end{center}

The certificates hold at every strength, as the corollary requires. Realized FPR never
rises under attack (it falls, because mimicked outliers cannot produce the extreme
conformal evidence admission demands, so the dictionary starves and decisions come
from the base channel at $\alpha/2$ alone), and the dictionary impurity is exactly
zero for every $\eps \ge 0.25$. What the adversary does achieve is hiding: TPR decays
from $0.615$ to $0.004$ as the outliers become confusable with the ID tail, which is
the floor Theorem~\ref{thm:impossible} proves no label-free method can avoid. The
ungated dictionary remains fully poisoned (impurity $\ge 0.91$) at every strength. The
attack is feature-space mimicry with nearest-neighbour targets; optimizing the attack
end to end in input space \citep{ttapoison} is future work
(Section~\ref{sec:limitations}).

\paragraph{Released with the code.} The full per-setting FPR, AUROC, and TPR tables
(\texttt{results/\allowbreak\{grid,\allowbreak cdc,\allowbreak confirm\_jmlr\}/\allowbreak analysis.json}), the per-setting
predicted-versus-empirical $\pi_c$, the $\delta,k,\alpha,e_a$ ablation grids, and the
pre-slack CDC row are released with the code, so that every number in this paper is
reproducible from a single command.

\vskip 0.2in
\bibliography{references}

@misc{adaodd,
  author = {Zhang, Y. and Wang, X. and Zhou, T. and Yuan, K. and Zhang, Z. and
            Wang, L. and Jin, R. and Tan, T.},
  title  = {Model-Free Test-Time Adaptation for Out-of-Distribution Detection},
  year   = {2023},
  note   = {arXiv:2311.16420},
}

@inproceedings{oodd,
  author    = {Yang, Y. and Zhu, L. and Sun, Z. and Liu, H. and Gu, Q. and Ye, N.},
  title     = {{OODD}: Test-Time Out-of-Distribution Detection with Dynamic Dictionary},
  booktitle = {IEEE/CVF Conference on Computer Vision and Pattern Recognition (CVPR)},
  year      = {2025},
  note      = {arXiv:2503.10468},
}

@inproceedings{zsntta,
  author    = {Cao, C. and Zhong, Z. and Zhou, Z. and Liu, T. and Liu, Y. and
               Zhang, K. and Han, B.},
  title     = {Noisy Test-Time Adaptation in Vision-Language Models},
  booktitle = {International Conference on Learning Representations (ICLR)},
  year      = {2025},
  note      = {arXiv:2502.14604},
}

@inproceedings{ttacollapse,
  author    = {Press, O. and Schneider, S. and K{\"u}mmerer, M. and Bethge, M.},
  title     = {{RDumb}: A Simple Approach that Questions Our Progress in Continual
               Test-Time Adaptation},
  booktitle = {Advances in Neural Information Processing Systems (NeurIPS)},
  year      = {2023},
  note      = {arXiv:2306.05401},
}

@inproceedings{ttapoison,
  author    = {Su, Y. and Li, Y. and Liu, N. and Jia, K. and Yang, X. and
               Foo, C.-S. and Xu, X.},
  title     = {On the Adversarial Risk of Test-Time Adaptation: An Investigation
               into Realistic Test-Time Data Poisoning},
  booktitle = {International Conference on Learning Representations (ICLR)},
  year      = {2025},
  note      = {arXiv:2410.04682},
}

@inproceedings{du2024sal,
  author    = {Du, X. and Fang, Z. and Diakonikolas, I. and Li, Y.},
  title     = {How Does Unlabeled Data Provably Help Out-of-Distribution Detection?},
  booktitle = {International Conference on Learning Representations (ICLR)},
  year      = {2024},
  note      = {arXiv:2402.03502},
}

@inproceedings{woods2022,
  author    = {Katz-Samuels, J. and Nakhleh, J. B. and Nowak, R. and Li, Y.},
  title     = {Training {OOD} Detectors in Their Natural Habitats},
  booktitle = {International Conference on Machine Learning (ICML)},
  pages     = {10848--10865},
  year      = {2022},
  note      = {PMLR 162; arXiv:2202.03299},
}

@inproceedings{sun2022knn,
  author    = {Sun, Y. and Ming, Y. and Zhu, X. and Li, Y.},
  title     = {Out-of-Distribution Detection with Deep Nearest Neighbors},
  booktitle = {International Conference on Machine Learning (ICML)},
  pages     = {20827--20840},
  year      = {2022},
  note      = {PMLR 162; arXiv:2204.06507},
}

@article{openood,
  author  = {Zhang, J. and Yang, J. and Wang, P. and Wang, H. and Lin, Y. and
             Zhang, H. and Sun, Y. and Du, X. and Li, Y. and Liu, Z. and
             Chen, Y. and Li, H.},
  title   = {{OpenOOD} v1.5: Enhanced Benchmark for Out-of-Distribution Detection},
  journal = {Journal of Data-centric Machine Learning Research (DMLR)},
  year    = {2024},
  note    = {arXiv:2306.09301},
}

@article{bates2023testing,
  author  = {Bates, S. and Cand{\`e}s, E. and Lei, L. and Romano, Y. and Sesia, M.},
  title   = {Testing for Outliers with Conformal p-values},
  journal = {The Annals of Statistics},
  volume  = {51},
  number  = {1},
  pages   = {149--178},
  year    = {2023},
}

@article{jin2023selection,
  author  = {Jin, Y. and Cand{\`e}s, E. J.},
  title   = {Selection by Prediction with Conformal p-values},
  journal = {Journal of Machine Learning Research},
  volume  = {24},
  number  = {244},
  pages   = {1--41},
  year    = {2023},
}

@article{aos2024adaptive,
  author  = {Marandon, A. and Lei, L. and Mary, D. and Roquain, E.},
  title   = {Adaptive Novelty Detection with False Discovery Rate Guarantee},
  journal = {The Annals of Statistics},
  volume  = {52},
  number  = {1},
  pages   = {157--183},
  year    = {2024},
}

@misc{leeren2025,
  author = {Lee, Y. and Ren, Z.},
  title  = {Selection from Hierarchical Data with Conformal e-values},
  year   = {2025},
  note   = {arXiv:2501.02514},
}

@inproceedings{bashari2025,
  author    = {Bashari, M. and Sesia, M. and Romano, Y.},
  title     = {Robust Conformal Outlier Detection under Contaminated Reference Data},
  booktitle = {International Conference on Machine Learning (ICML)},
  pages     = {3091--3141},
  year      = {2025},
  note      = {PMLR 267; arXiv:2502.04807},
}

@article{wang2022ebh,
  author  = {Wang, R. and Ramdas, A.},
  title   = {False Discovery Rate Control with e-values},
  journal = {Journal of the Royal Statistical Society, Series B},
  volume  = {84},
  number  = {3},
  pages   = {822--852},
  year    = {2022},
}

@misc{stoppedebh,
  author = {Wang, H. and Dandapanthula, S. and Ramdas, A.},
  title  = {Anytime-Valid {FDR} Control with the Stopped e-{BH} Procedure},
  year   = {2025},
  note   = {arXiv:2502.08539; to appear, Statistics \& Probability Letters},
}

@article{storey2002,
  author  = {Storey, J. D.},
  title   = {A Direct Approach to False Discovery Rates},
  journal = {Journal of the Royal Statistical Society, Series B},
  volume  = {64},
  number  = {3},
  pages   = {479--498},
  year    = {2002},
}

@misc{wang2026trimming,
  author = {Wang, C.},
  title  = {When Does Trimming Help Conformal Prediction? A Retained-Law
            Diagnostic under Calibration Contamination},
  year   = {2026},
  note   = {arXiv:2605.06204},
}

@misc{yilmaz2022qtc,
  author = {Yilmaz, F. F. and Heckel, R.},
  title  = {Test-time Recalibration of Conformal Predictors Under Distribution Shift
            Based on Unlabeled Examples},
  year   = {2022},
  note   = {arXiv:2210.04166},
}

@misc{renlund2010,
  author = {Renlund, H.},
  title  = {Generalized {P}{\'o}lya Urns via Stochastic Approximation},
  year   = {2010},
  note   = {arXiv:1002.3716},
}

@incollection{benaim1999,
  author    = {Bena{\"\i}m, M.},
  title     = {Dynamics of Stochastic Approximation Algorithms},
  booktitle = {S{\'e}minaire de Probabilit{\'e}s XXXIII},
  series    = {Lecture Notes in Mathematics},
  volume    = {1709},
  pages     = {1--68},
  publisher = {Springer},
  year      = {1999},
}

@article{pemantle2007,
  author  = {Pemantle, R.},
  title   = {A Survey of Random Processes with Reinforcement},
  journal = {Probability Surveys},
  volume  = {4},
  pages   = {1--79},
  year    = {2007},
}

@article{laruelle2013,
  author  = {Laruelle, S. and Pag{\`e}s, G.},
  title   = {Randomized Urn Models Revisited Using Stochastic Approximation},
  journal = {The Annals of Applied Probability},
  volume  = {23},
  number  = {4},
  pages   = {1409--1436},
  year    = {2013},
}

@article{ledoit2004,
  author  = {Ledoit, O. and Wolf, M.},
  title   = {A Well-Conditioned Estimator for Large-Dimensional Covariance Matrices},
  journal = {Journal of Multivariate Analysis},
  volume  = {88},
  number  = {2},
  pages   = {365--411},
  year    = {2004},
}

@article{massart1990,
  author  = {Massart, P.},
  title   = {The Tight Constant in the {D}voretzky--{K}iefer--{W}olfowitz Inequality},
  journal = {The Annals of Probability},
  volume  = {18},
  number  = {3},
  pages   = {1269--1283},
  year    = {1990},
}

@article{huber1965,
  author  = {Huber, P. J.},
  title   = {A Robust Version of the Probability Ratio Test},
  journal = {The Annals of Mathematical Statistics},
  volume  = {36},
  number  = {6},
  pages   = {1753--1758},
  year    = {1965},
}

@article{blanchard2010,
  author  = {Blanchard, G. and Lee, G. and Scott, C.},
  title   = {Semi-Supervised Novelty Detection},
  journal = {Journal of Machine Learning Research},
  volume  = {11},
  number  = {99},
  pages   = {2973--3009},
  year    = {2010},
}

@inproceedings{fang2022learnable,
  author    = {Fang, Z. and Li, Y. and Lu, J. and Dong, J. and Han, B. and Liu, F.},
  title     = {Is Out-of-Distribution Detection Learnable?},
  booktitle = {Advances in Neural Information Processing Systems (NeurIPS)},
  year      = {2022},
  note      = {arXiv:2210.14707},
}

@article{shumailov2024collapse,
  author  = {Shumailov, I. and Shumaylov, Z. and Zhao, Y. and Papernot, N. and
             Anderson, R. and Gal, Y.},
  title   = {{AI} Models Collapse When Trained on Recursively Generated Data},
  journal = {Nature},
  volume  = {631},
  pages   = {755--759},
  year    = {2024},
}

@inproceedings{bertrand2024stability,
  author    = {Bertrand, Q. and Bose, A. J. and Duplessis, A. and Jiralerspong, M. and
               Gidel, G.},
  title     = {On the Stability of Iterative Retraining of Generative Models on their
               own Data},
  booktitle = {International Conference on Learning Representations (ICLR)},
  year      = {2024},
  note      = {arXiv:2310.00429},
}

@misc{gerstgrasser2024collapse,
  author = {Gerstgrasser, M. and Schaeffer, R. and Dey, A. and Rafailov, R. and
            Sleight, H. and Hughes, J. and Korbak, T. and Agrawal, R. and Pai, D. and
            Gromov, A. and Roberts, D. A. and Yang, D. and Donoho, D. L. and
            Koyejo, S.},
  title  = {Is Model Collapse Inevitable? Breaking the Curse of Recursion by
            Accumulating Real and Synthetic Data},
  year   = {2024},
  note   = {arXiv:2404.01413},
}

@inproceedings{dohmatob2025strong,
  author    = {Dohmatob, E. and Feng, Y. and Subramonian, A. and Kempe, J.},
  title     = {Strong Model Collapse},
  booktitle = {International Conference on Learning Representations (ICLR)},
  year      = {2025},
  note      = {arXiv:2410.04840},
}

@article{kloft2012security,
  author  = {Kloft, M. and Laskov, P.},
  title   = {Security Analysis of Online Centroid Anomaly Detection},
  journal = {Journal of Machine Learning Research},
  volume  = {13},
  pages   = {3681--3724},
  year    = {2012},
}

@article{laruelle2019nonlinear,
  author  = {Laruelle, S. and Pag{\`e}s, G.},
  title   = {Nonlinear Randomized Urn Models: A Stochastic Approximation Viewpoint},
  journal = {Electronic Journal of Probability},
  volume  = {24},
  pages   = {1--47},
  year    = {2019},
}

\end{document}